\documentclass[9.5pt,cspaper,compsoc,final]{IEEEtran}
\usepackage{packages}
\usepackage{customized}
\usepackage{ragged2e}
\usepackage{etoolbox}

\providetoggle{final}
\settoggle{final}{false}
\iftoggle{final}{
\hypersetup{
    pagebackref=false,
    breaklinks=true,
    bookmarks=false,
    linkcolor=black,
    citecolor=black,
    urlcolor=black,
    }
}{

}

\def\jian{\textcolor{black}}
\def\DingJian{\textcolor{black}}

\begin{document}
	\title{Object Detection in Aerial Images: \\ A Large-Scale Benchmark and Challenges}
	\author{
		Jian~Ding, 
		\and Nan~Xue,
		\and Gui-Song~Xia, 
		\and Xiang~Bai, 
		\and Wen Yang, 
		\and Michael Ying~Yang, \\
		\and Serge~Belongie, 
		\and Jiebo~Luo, 
		\and Mihai~Datcu, 
		\and Marcello~Pelillo, 
	\and	Liangpei~Zhang
		\IEEEcompsocitemizethanks{
			\IEEEcompsocthanksitem J. Ding and L. Zhang are with the State Key Lab. LIESMARS, Wuhan University, Wuhan, 430079, China.  Email: \{{jian.ding, zlp62}\}@whu.edu.cn.
			\IEEEcompsocthanksitem N. Xue is with the National Engineering Research Center for Multimedia Software, School of Computer Science and Institute of Artificial Intelligence, Wuhan University, Wuhan, 430072, China.  Email: xuenan@whu.edu.cn.
			\IEEEcompsocthanksitem G.-S. Xia is with the National Engineering Research Center for Multimedia Software, School of Computer Science and Institute of Artificial Intelligence, and also the State Key Lab. LIESMARS, Wuhan University, Wuhan, 430072, China. Email: {guisong.xia}@whu.edu.cn.
			\IEEEcompsocthanksitem X. Bai is with the School of Electronic Information, Huazhong University of Science and Technology, Wuhan, 430079, China. Email: xbai@hust.edu.cn.
			\IEEEcompsocthanksitem W. Yang is with the School of Electronic Information, and the State Key Lab. LIESMARS, Wuhan University, Wuhan, 430072, China.  Email: yangwen@whu.edu.cn.
			\IEEEcompsocthanksitem M. Y.~Yang is with Faculty of Geo-Information Science and Earth Observation (ITC), University of Twente, the Netherlands. Email: michael.yang@utwente.nl.
			\IEEEcompsocthanksitem S. Belongie is with Department of Computer Science, Cornell University and Cornell Tech. Email: sjb344@cornell.edu.
			\IEEEcompsocthanksitem J. Luo is with Department of Computer Science, University of Rochester, Rochester, NY 14627. Email: jluo@cs.rochester.edu.
			\IEEEcompsocthanksitem M. Datcu is with Remote Sensing Technology Institute, German Aerospace Center (DLR), 82234, Germany, and also the University POLITEHNICA of Bucharest (UPB), Romania. Email: mihai.datcu@dlr.de.
			\IEEEcompsocthanksitem M. Pelillo is with DAIS, Ca' Foscari University of Venice, Italy. Email: pelillo@unive.it.
			\IEEEcompsocthanksitem The studies in this paper have been supported by the NSFC projects under the contracts No.61922065, No.61771350 and No.41820104006. Dr. Nan Xue was also supported by National Post-Doctoral Program for Innovative Talents under Grant BX20200248. Mihai Datcu was supported by the CNCS-UEFISCDI, project number PN-III-P4-ID-PCE-2020-2120, within PNCDI III. 
			\IEEEcompsocthanksitem The corresponding author is Gui-Song Xia (guisong.xia@whu.edu.cn).
		}
	}

	\IEEEtitleabstractindextext{%
		\justify
		\begin{abstract}
			In the past decade, object detection has achieved significant progress in natural images but not in aerial images, due to the massive variations in the scale and orientation of objects caused by the bird's-eye view of aerial images. More importantly, the lack of large-scale benchmarks has become a major obstacle to the development of object detection in aerial images (ODAI).
			In this paper, we present a large-scale {\em Dataset of Object deTection in Aerial images} (DOTA) and comprehensive baselines for ODAI. The proposed DOTA dataset contains 1,793,658 object instances of 18 categories of oriented-bounding-box annotations collected from 11,268 aerial images. Based on this large-scale and well-annotated dataset, we build baselines covering 10 state-of-the-art algorithms with over 70 configurations, where the speed and accuracy performances of each model have been evaluated. Furthermore, we provide a code library for ODAI and build a website for evaluating different algorithms. Previous challenges run on DOTA have attracted more than 1300 teams worldwide. We believe that the expanded large-scale DOTA dataset, the extensive baselines, the code library and the challenges can facilitate the designs of robust algorithms and reproducible research on the problem of object detection in aerial images.			
		\end{abstract}		
}
\maketitle

\begin{figure*}[t!]
			\centering
			\includegraphics[width=0.87\linewidth]{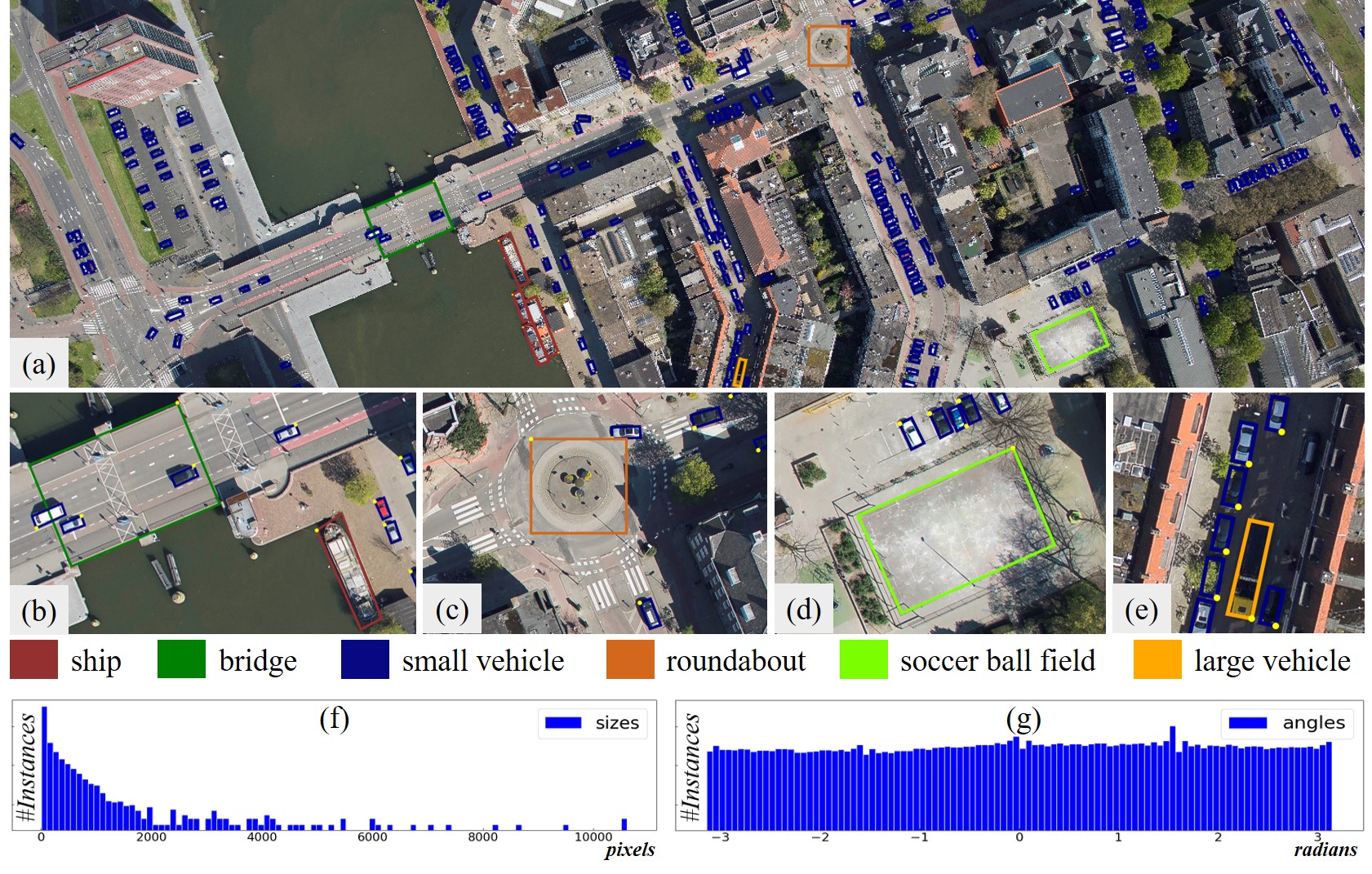}
			\vspace{-4mm}
			\caption{{\bf An example image taken from DOTA}. 
				(a) A typical image in DOTA consisting of many instances from multiple categories. (b), (c), (d), (e) are cropped from the source image. We can see that instances such as small vehicles have arbitrary orientations. There is also a massive scale variation across different instances. Moreover, the instances are not distributed uniformly. The instances are sparse in most areas but crowded in some local areas. Large vehicles and ships have large ARs. (f) and (g) exhibit the size and orientation histograms, respectively, for all instances.
			}
			\vspace{-3mm}
			\label{fig:large-example}
		\end{figure*}

\IEEEraisesectionheading{
\section{Introduction}\label{sec:introduction}
}	
        Currently, Earth vision (also known as Earth observation and remote sensing) technologies enable us to observe the Earth`s surface with aerial images\footnote{This paper uses the term "aerial" to refer to any overhead image looking approximately straight down onto the Earth, including both satellite images and airborne images, for simplification unless otherwise indicated. We use the term ``airborne" if we do not want to include the satellite images.} with a resolution up to a half meter. Although challenging, developing mathematical tools and numerical algorithms is necessary for interpreting these huge volumes of images, among which object detection refers to localizing objects of interest (\eg, vehicles and ships) on the Earth`s surface and predicting their categories. Object detection in aerial images (ODAI) has been an essential step in many real-world applications such as urban management, precision agriculture, emergency rescue and disaster relief~\cite{argriculture, surveillance}. Although extensive studies have been devoted to object detection in aerial images and appreciable breakthroughs have been made~\cite{LHI_method, HRSC2016,Detect_car_in_uav,RIFD,SZTAKI-INRIA,CARPPK}, the task has numerous difficulties such as arbitrary orientations, scale variations, extremely nonuniform object densities and large aspect ratios (ARs), as shown in Fig.~\ref{fig:large-example}.  
		
		Among these difficulties, the arbitrary orientation of objects caused by the overhead view is the main difference between natural images and aerial images, and it complicates the object detection task in two ways. First, rotation-invariant feature representations are preferred in the detection of arbitrarily orientated objects, but they are often beyond the capability of most of current deep neural network models. Although the methods such as those designed in~\cite{RICNN,RIFD,ORN} use rotation-invariant convolutional neural networks (CNNs), the problem is far from solved. Second, the {\em horizontal bounding box} (HBB) object representation used in conventional object detection~\cite{PASCALVOC, Imagenet,COCO} cannot localize the oriented objects precisely, such as ships and large vehicles, as shown in Fig.~\ref{fig:large-example}. The {\em oriented bounding box} (OBB) object representation is more appropriate for aerial images~\cite{DOTA, VEDAI, DLR3KMunichVehicle, HRSC2016, ucas-aod}. It allows us to distinguish densely packed instances (as shown in Fig.~\ref{fig:compare_annotations}) and extract rotation-invariant features~\cite{HRSC2016,RoITransformer,ICN}. The OBB object representation actually introduces a new object detection task, called {\em oriented object detection}. 
		In contrast with horizontal object detection~\cite{Fast_Tiny, clusternet, Clustered_Aerial, CARPPK}, oriented object detection is a recently emerging research direction and most of the methods for this new task attempt to transfer successful deep-learning-based object detectors pre-trained on large-scale natural image datasets (\eg, ImageNet~\cite{Imagenet} and Microsoft Common Objects in Context (MS COCO)~\cite{ COCO }) to aerial scenes~\cite{ICN, RoITransformer,RRD,scrdet,li2019learning} due to the lack of large-scale annotated aerial image datasets.

		To mitigate the dataset problem, some public datasets of aerial images have been created, see \eg~\cite{TAS, SZTAKI-INRIA, VEDAI, DLR3KMunichVehicle, ucas-aod}, but they contain a limited number of instances and tend to use images taken under ideal conditions (\eg, clear backgrounds and centered objects), which cannot reflect the real-world difficulties of the problem. The recently released xView~\cite{xview} dataset provides a wide range of categories and contains large quantities of instances in complicated scenes. However, it annotates the instances with HBBs instead of the more precise OBBs. Thus, a large-scale dataset that has OBB annotations and reflects the difficulties in real-world applications of aerial images is in high demand.

		Another issue with ODAI is that the module design and the hyperparameter setting of conventional object detectors learned from natural images are not appropriate for aerial images due to domain differences. Thus, when developing new algorithms, comprehensive baselines and sufficient ablative analyses of models on aerial images are required. However, comparing different algorithms is difficult due to the diversities in hardware, software platforms, detailed settings and so on. These factors influence both speed and accuracy. Therefore, when building the baselines, implementing the algorithms with a unified code library and keeping the hardware and software platform the same is highly desirable. Nevertheless, current object detection libraries, \eg, MMDetection~\cite{mmdetection} and Detectron~\cite{Detectron2018}, do not support oriented object detection.		
		
		To address the above-mentioned problems, in this paper we first extend the preliminary version of DOTA, \ie,  DOTA-v1.0~\cite{DOTA}, to DOTA-v2.0. Specifically, DOTA-v2.0 collects $11,268$ aerial images from various sensors and platforms and contains {\bf approximately 1.8 million object instances} annotated with {\bf OBBs} in $18$ common categories, which, to our knowledge, is the largest public Earth vision object detection dataset. 
		Then, to facilitate algorithm developments and comparisons with DOTA, we provide a well-designed code library that supports oriented object detection in aerial images. Based on the code library, we also build more comprehensive baselines than the preliminary version~\cite{DOTA}, keeping the hardware, software platform, and settings the same. In total, we evaluate $10$ algorithms and over $70$ models with different configurations. We then provide detailed speed and accuracy analyses to explore the module designs and parameter settings in aerial images to guide future research.
		These experiments verify the large differences in object detector design between natural and aerial images and provide materials for universal object detection algorithms~\cite{universalobject}. 
		
		The main contributions of this paper are three-fold:
		\begin{itemize}
		  \vspace{-1mm}
			\item 
			To the best of our knowledge, the expanded DOTA is the largest dataset for object detection in Earth vision. The OBB annotations of DOTA not only provide a large-scale benchmark for object detection in Earth vision but also pose interesting algorithmic questions and challenges to generalized object detection in computer vision.

			\item
			We build a code library for object detection in aerial images.  This is expected to facilitate the development and benchmarking of object detection algorithms in aerial images with both HBB and OBB representations. 
			
			\item
			With the expanded DOTA, we evaluate $10$ representative algorithms over $70$ model configurations, providing comprehensive analysis that can guide the designs of object detection algorithms in aerial images.
		\end{itemize}
		
The dataset, code library, and regular evaluation server are available and maintained on the DOTA website\footnote{\url{https://captain-whu.github.io/DOTA/}}. 
It is worth noting that the creation and use of DOTA have advanced object detection in aerial images. For instance, the regular DOTA evaluation server and two object detection contests organized at the 2018 International Conference on Pattern Recognition (ICPR'~2018 with DOTA-v1.0)\footnote{\url{https://captain-whu.github.io/ODAI/results.html}} and 2019 Conference on Computer Vision and Pattern Recognition (CVPR'2019 with DOTA-v1.5)\footnote{\url{https://captain-whu.github.io/DOAI2019/challenge.html}} have attracted approximately $1300$ registrations. 
We believe that our new DOTA dataset, with a comprehensive code library and an online evaluation platform, will further promote the reproducible research in Earth vision.

		
		\section{Related work}
		
		Well-annotated datasets have played an important role in data-driven computer vision research~\cite{LHI_CVPR,Imagenet,COCO,Places,EmotionRecognition, AID,openimage-v4} and have promoted cutting-edge research in a number of tasks such as object detection and classification. In this section, we first review object detection datasets of natural and aerial images. Then we discuss the recent deep learning based object detectors in aerial images. Finally, we briefly review the code libraries for object detection.

		\subsection{Datasets for Conventional Object Detection}
		As a pioneer, PASCAL Visual Object Classes (VOC)~\cite{PASCALVOC} held challenges on object detection from 2005 to 2012. The computer vision community has widely adopted PASCAL VOC datasets and their evaluation metrics. Specifically, the PASCAL VOC Challenge 2012 dataset contains $11,530$ images, $20$ classes, and $27,450$ annotated bounding boxes. Later, the ImageNet dataset~\cite{Imagenet} was developed and is an order of magnitude larger than PASCAL VOC, containing $200$ classes and approximately $500,000$ annotated bounding boxes. However, non-iconic views are not addressed. Then MS COCO~\cite{COCO} was released, containing a total of $328$K images, $91$ categories, and $2.5$ million labeled segmented objects. MS COCO has on average more instances and categories per image and contains more contextual information than PASCAL VOC and ImageNet. 
		It is worth noticing that, in Earth vision, the image size could be extremely large (\eg, $20,000\times20,000$ pixels), so the number of images cannot reflect the scale of a dataset. In this case, the pixel area would be more reasonable when comparing the scale between the datasets of natural and aerial images. Moreover, the large images include more instances per image and contextual information. Tab.~\ref{compare_general_dataset} provides the detailed comparisons.
    	\begin{table}[htp!]
		    \centering
			\footnotesize
			\caption{DOTA {\em vs.} general object detection datasets. {\em BBox} is short for bounding box, {\em Avg. BBox quantity} indicates the average number of bounding boxes per image. For PASCAL VOC (07++12), we count the whole PASCAL VOC 07 and training and validation (trainval) set of PASCAL VOC 12. DOTA has a comparable scale with the large-scale datasets for object detection in natural images. Note that for the average number of instances per image, DOTA surpasses the other datasets.}
			\vspace{-3mm}
			\setlength{\tabcolsep}{1.3mm}
			\resizebox{0.95\linewidth}{!}{
			\begin{tabular}{cccccc}
				\hline
				Dataset                                                           & Classes & \begin{tabular}[c]{@{}c@{}}Image \\ quantity \end{tabular} & \begin{tabular}[c]{@{}c@{}}Megapixel \\ area\end{tabular} & \begin{tabular}[c]{@{}c@{}}BBox \\ quantity \end{tabular} & \begin{tabular}[c]{@{}c@{}}Avg. BBox \\ quantity \end{tabular} \\ \hline
				\begin{tabular}[c]{@{}c@{}}PASCAL VOC \\ (07++12)\end{tabular}    & 20      & 21,503                                                     & 5,133                                                     & 52,090                                                    & 2.42                                                           \\ \hline
				\begin{tabular}[c]{@{}c@{}}MS COCO \\ (2014 trainval)\end{tabular} & 80      & 123,287                                                    & 32,639                                                    & 886,266                                                   & 7.19                                                           \\ \hline
				\begin{tabular}[c]{@{}c@{}}ImageNet \\ (2014 train)\end{tabular}  & 200     & 456,567                                                    & 82,820                                                    & 478,807                                                   & 1.05                                                           \\ \hline
				DOTA-v1.0                                                         & 15      & 2,806                                                      & 19,173                                                    & 188,282                                                   & 67.10                                                          \\ \hline
				DOTA-v1.5                                                         & 16      & 2,806                                                       & 19,173                                                     & 402,089                                                   & 143.73                                                         \\ \hline
				DOTA-v2.0                                                         & 18      & 11,268                                                     & \textbf{126,306}                                          & \textbf{1,793,658}                                        & \textbf{159.18}                                                \\ \hline
			\end{tabular}
			}
			\vspace{-3mm}
			\label{compare_general_dataset}		
		\end{table}
	
		\subsection{Datasets for Object Detection in Aerial Images}  
		In aerial object detection, a dataset resembling MS COCO and ImageNet both in terms of the image number and detailed annotations has been missing, which becomes one of the main obstacles to research in Earth vision, especially for developing deep-learning-based algorithms. 
		In Earth vision, many aerial image datasets are prepared for actual demands in a specific category, such as building datasets~\cite{SZTAKI-INRIA, SpaceNet_MVOI}, vehicle datasets~\cite{TAS, VEDAI, COWC, DLR3KMunichVehicle, CARPPK, ITCVD, yangITCVD19}, ship datasets~\cite{HRSC2016,chen2020fgsd}, and plane datasets~\cite{shermeyer2020rareplanes,ucas-aod}. 
		Although some public datasets~\cite{ucas-aod, VHR, RSOD, LEVIR, HRRSD} have multiple categories, they have only limited number of samples, which are hardly efficient for training robust deep models. For example, NWPU~\cite{VHR} only contains $800$ images, $10$ classes and $3,651$ instances.
		 
        To alleviate this problem, our preliminary work DOTA-v1.0~\cite{DOTA} presented a dataset with $15$ categories and $188,282$ instances, which for the first time enables us to efficiently train robust deep models for ODAI without the help of large-scale datasets of natural images, such as MS COCO and ImageNet. Later, iSAID~\cite{iSAID} provided an instance segmentation extension of DOTA-v1.0~\cite{DOTA}. 
        A notable dataset is xView~\cite{xview}, which contains $1,413$ images, $16$ main categories, $60$ fine-grained categories, and $1$ million instances. Another dataset DIOR~\cite{DIOR} provided a comparable number of instances as DOTA-v1.0~\cite{DOTA}. However, the instances in xView and DIOR are both annotated by HBBs, which are not suitable for precisely detecting objects that are arbitrarily oriented in aerial images. In addition, VisDrone~\cite{visdrone} is also a large-scale dataset for drone images but focuses more on video object detection and tracking. The image subset in VisDrone for object detection is not very large. Furthermore, most of the previous datasets are heavily imbalanced in favor of positive samples, whose negative samples are not sufficient to represent the real-world distribution. 
		\begin{table*}[htb!]
			\small
			\centering
			\caption{ DOTA vs. object detection datasets in aerial images. \textit{HBB} is {\em horizontal bounding box}, and \textit{OBB} is {\em oriented bounding box} . \textit{CP} is {\em center point}.}
			\vspace{-3mm}
			\resizebox{0.95\linewidth}{!}{
			\begin{tabular}{c|cccccccc}
				\hline
				Dataset                            & Source       & Annotation & \begin{tabular}[c]{@{}c@{}}\# of main \\ categories\end{tabular} & \begin{tabular}[c]{@{}c@{}}Total \# of \\ categories\end{tabular} & \# of instances & \# of images & \begin{tabular}[c]{@{}c@{}}Image \\ width\end{tabular} & Year                     \\ \hline
				TAS~\cite{TAS}                     & satellite    & HBB        & 1                                                            & 1                                                              & 1,319       & 30       & 792                                                    & 2008                     \\
				SZTAKI-INRIA~\cite{SZTAKI-INRIA}   & multi source & OBB        & 1                                                            & 1                                                              & 665         & 9        & $\sim$800                                              & 2012                     \\
				NWPU VHR-10~\cite{VHR}             & multi source & HBB        & 10                                                           & 10                                                             & 3,651       & 800      & $\sim$1000                                             & 2014                     \\
				VEDAI~\cite{VEDAI}                 & satellite    & OBB        & 3                                                            & 9                                                              & 2,950       & 1,268    & 512, 1024                                              & 2015                     \\
				DLR 3k~\cite{DLR3KMunichVehicle}   & aerial       & OBB        & 2                                                            & 8                                                              & 14,235      & 20       & 5616                                                   & 2015                     \\
				UCAS-AOD~\cite{ucas-aod}           & Google Earth & OBB        & 2                                                            & 2                                                              & 14,596      & 1,510    & $\sim$1000                                             & 2015                     \\
				COWC~\cite{COWC}                   & aerial       & CP         & 1                                                            & 1                                                              & 32,716      & 53       & 2000$-$19,000                                          & 2016                     \\
				HRSC2016~\cite{HRSC2016}           & Google Earth & OBB        & 1                                                            & 26                                                             & 2,976       & 1,061    & $\sim$1100                                             & 2016                     \\
				RSOD~\cite{RSOD}                   & Google Earth & HBB        & 4                                                            & 4                                                              & 6,950       & 976      & $\sim$1000                                             & \multicolumn{1}{l}{2017} \\
				CARPPK~\cite{CARPPK}               & drone        & HBB        & 1                                                            & 1                                                              & 89,777      & 1,448    & 1280                                                   & \multicolumn{1}{l}{2017} \\
				ITCVD~\cite{ITCVD}                 & aerial       & HBB        & 1                                                            & 1                                                              & 228         & 23,543   & 5616                                                   & \multicolumn{1}{l}{2018} \\
				LEVIR~\cite{LEVIR}                 & Google Earth & HBB        & 3                                                            & 3                                                              & 11,000      & 22,000   & 800$-$600                                              & \multicolumn{1}{l}{2018} \\
				xView~\cite{xview}                 & satellite    & HBB        & 16                                                           & 60                                                             & 1,000,000   & 1,413    & $\sim$3000                                             & \multicolumn{1}{l}{2018} \\
				VisDrone~\cite{visdrone}           & drone        & HBB        & 10                                                           & 10                                                             & 54,200      & 10,209   & 2000                                                   & \multicolumn{1}{l}{2018} \\
				SpaceNet MVOI~\cite{SpaceNet_MVOI} & satellite    & polygon    & 1                                                            & 1                                                              & 126,747     & 60,000   & 900                                                    & \multicolumn{1}{l}{2019} \\
				HRRSD~\cite{HRRSD}                 & multi source & HBB        & 13                                                           & 13                                                             & 55,740      & 21,761   & 152$-$10569                                            & 2019                     \\
				DIOR~\cite{DIOR}                   & Google Earth & HBB        & 20                                                           & 20                                                             & 190,288     & 23,463   & 800                                                    & \multicolumn{1}{l}{2019} \\
				iSAID~\cite{iSAID}                 & multi source & polygon        & 14                                                           & 15                                                             & 655,451     & 2,806    & 800$-$13,000                                           & \multicolumn{1}{l}{2019} \\
				FGSD~\cite{chen2020fgsd}	                   & Google Earth &	OBB   &	1 &	43 
				                          & 5,634 & 2,612 & 930 &  \multicolumn{1}{l}{2020} \\
                RarePlanes~\cite{shermeyer2020rareplanes} &	satellite & polygon & 1 & 110 & 644,258 & 50,253 &	1080 & \multicolumn{1}{l}{2020} \\

				 \hline
				DOTA-v1.0~\cite{DOTA}              & multi source & OBB        & 14                                                           & 15                                                             & 188,282     & 2,806     & 800$-$13,000                                           & 2018                     \\
				DOTA-v1.5             & multi source & OBB        & 15                                                           & 16                                                             & 402,089     & 2,806     & 800$-$13,000                                           & 2019                     \\
				DOTA-v2.0                          & multi source & OBB        & 17                                                           & 18                                                             & 1,793,658   & 11,268   & 800$-$20,000                                           & 2021                     \\ \hline
			\end{tabular}
			}
			\vspace{-3mm}
			\label{former-datasets}
		\end{table*}

		As we stated previously~\cite{DOTA}, a good dataset for aerial image object detection should have the following properties: 1) substantial annotated data to facilitate data-driven, especially deep-learning-based methods; 2) large images to contain more contextual information; 3) OBB annotation to describe the precise location of objects; and 4) balance in image sources, as pointed in~\cite{torralba2011unbiased}. DOTA is built considering these principles (unless otherwise specified, DOTA refers to DOTA-v2.0). Detailed comparisons of these existing datasets and DOTA are shown in Tab.~\ref{former-datasets}.
		Compared to other aerial datasets, as we shall see in Sec.~\ref{statistics_DOTA}, DOTA is challenging due to its large number of object instances, arbitrary orientations, various categories, density distribution, and diverse aerial scenes from various image sources. These properties make DOTA helpful for real-world applications.

			\begin{figure*}
				\centering
				\includegraphics[width=0.78\linewidth]{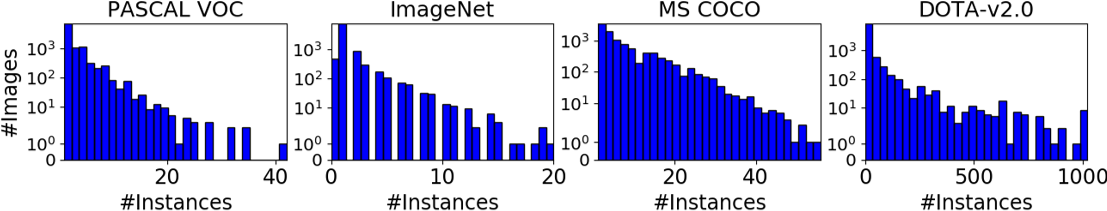}
				\vspace{-4mm}
				\caption{{\bf Number of instances per image among DOTA and general object detection datasets.} For PASCAL, ImageNet and MS COCO, we count the statistics of 10,000 random images. As the images in DOTA are very large ($20,000\times20,000$), for a fair comparison, we count the statistics of 10,1000 image patches
with the size of $1024\times1024$, which is also the size used for the baselines in Sec.~\ref{sec:benchmarks}. DOTA-v2.0 has a wider range of the number of instances per image.}
		\vspace{-3mm}
			\label{fig:instance_per_image_compare}
			\end{figure*}

			\subsection{Deep Models for Object Detection in Aerial Images}
			Object detection in aerial images is a longstanding problem.
			Recently, with the development of deep learning, many researchers in Earth vision have adapted deep object detectors~\cite{SSD,yolo9000, r-cnn, Fast_R-CNN, FasterR-CNN} developed for natural images to aerial images. However, the challenges caused by the domain shift need to be addressed. Here, we highlight some notable works.
			
            Objects in aerial images are often arbitrarily oriented due to the bird's-eye view, and the scale variations are larger than those in natural images. 
            To handle rotation variations, a simple model~\cite{RICNN} plugs an additional rotation-invariant layer into R-CNN~\cite{r-cnn} relying on rotation data augmentation. The oriented response network (ORN) introduces active rotating filters (ARF) to produce the rotation-invariant feature without using data augmentation, which is adopted by the rotation-sensitive regression detector (RRD)~\cite{RRD}. The deformable modules~\cite{Deformable} designed for general object deformation are also widely used in aerial images. The methods mentioned above do not fully utilize the OBB annotations. When OBB annotations are available, a rotation R-CNN (RR-CNN)~\cite{RR-CNN} uses rotation region-of-interest (RRoI) pooling to extract rotation-invariant region features. 
            However, RR-CNNs~\cite{RR-CNN} generate proposals by hand-crafted way.
            Then the RoI Transformer~\cite{RoITransformer} tries to use the supervision of OBBs to learn RoI-wise spatial transformation. The later S$^2$A-Net~\cite{han2020align}  extracts spatially invariant features in one-stage detectors. 
            To solve the challenges of scale variations, feature pyramids~\cite{FPN,ICN} and image pyramids~\cite{scrdet, li2019learning} are widely used to extract scale-invariant features in aerial images.
            We evaluate the geometric transformation network modules and geometric data augmentations in Sec.~\ref{sec:benchmarkresults}.

            Crowded instances represented by HBBs are difficult to distinguish (see Fig.~\ref{fig:compare_annotations}). Traditional HBB-based non maximum suppression (NMS) will fail in such cases. Therefore, these methods~\cite{RoITransformer,li2019learning,scrdet} use rotated NMS (R-NMS), which requires precise detections to address this problem. Similar to text and face detection in natural scenes, \eg~\cite{RRD,ITN,rotation_face,PCN}, precise ODAI can also be modeled as an oriented object detection task. Most of the previous works~\cite{RRD, textboxes++,DOTA, scrdet, li2019learning} consider it as a regression problem and regress the offsets of the OBB ground truth relative to anchors (or proposals). However, the definition of an OBB is ambiguous. For example, there are four permutations of the corner points in a quadrilateral. The Faster R-CNN OBB~\cite{DOTA} solves it by using a defined rule to determine the order of points in OBBs. Work in~\cite{Glidingertex} further uses the gliding offset and obliquity factor to eliminate the ambiguity. The circular smooth label (CSL)~\cite{Circular-Smooth-Label} transforms the regression of angle as a classification problem to avoid the problem. Mask OBB~\cite{maskobb} and CenterMap~\cite{CenterMap} consider object detection as a pixel-level classification problem to avoid ambiguity. Mask-based methods converge more easily but have more floating point operations per second (FLOPS) than regression-based methods. We will give a more detailed comparison between them in one unified code library in Sec.~\ref{sec:maskvsobb}.

            The final challenge is detecting objects in large images.
            Aerial images are usually extremely large (over $20k\times20k$ pixels). 
            Current GPU memory capacity is insufficient to process large images.
            Downsampling a large image to a small size would lose the detailed information. To solve this problem~\cite{DOTA,DLR3KMunichVehicle}, the large images can be simply split into small patches. After obtaining the results on these patches, the results are integrated into large images. 
To speed up inference on large images, these methods~\cite{Fast_Tiny,clusternet,Clustered_Aerial,uzkent2019efficient} first find regions that are likely to contain instances in the large images and then detect objects in the regions. In this paper, we simply follow the naive solutions~\cite{DOTA,DLR3KMunichVehicle} to build baselines.

\subsection{Code Libraries for Object Detection}

The development of object detection algorithms is a sophisticated process. In addition, there are too many design choices and hyperparameter settings, which make comparisons between different methods difficult. Therefore, object detection code libraries such as the Tensorflow Object Detection API~\cite{huang2016speed}, Detectron~\cite{Detectron2018}, MaskRCNN-Benchmark~\cite{maskrcnnbenchmark}, Detectron2~\cite{detectron2}, MMDetection~\cite{mmdetection} and SimpleDet\cite{simpledet} are developed to facilitate the comparisons of object detection algorithms. These code libraries primarily use a modular design, which makes it easy to develop new algorithms. The current widely used settings, such as the training schedule, are from Detectron~\cite{Detectron2018}.  However, these code libraries mainly focus on horizontal object detection. Only Detectron2~\cite{detectron2} has limited support for oriented object detection. In our work, we enriched MMDetection~\cite{mmdetection} with several crucial operators for oriented object detection and evaluated 10 algorithms for object detection in aerial images.

			\section{Construction of DOTA}
			\subsection{Image Collection}
			
			In aerial images, the resolution and a variety of sensors are the factors that produce dataset biases~\cite{earthvisonba}. To eliminate these biases, we collect images from various sensors and platforms with multiple resolutions, including Google Earth, the Gaofen-2 (GF-2) Satellite, Jilin-1 (JL-1) Satellite, and airborne images (taken by CycloMedia~\cite{cyclomedia} in Rotterdam). To obtain the DOTA images, we first collected the coordinates of areas of interest (\eg, airports or harbors) from all over the world. Then, according to the coordinates, images are collected from Google Earth, GF-2 and JL-1 (GF\&JL) satellites. The airborne images taken by CycloMedia~\cite{cyclomedia} were obtained from five perspectives in Rotterdam, which include both oblique views and nadir views. 
			The tilt angle of the oblique view was approximately $45^{\circ}$. 
			
			For the Google Earth images, we collect the images that contain instances of interest with sizes from $800 \times 800$ to $4000 \times 4000$ pixels. However, for the GF\&JL satellite and airborne images, we maintained their original sizes. Large images can approach real-world distributions, and also pose a challenge for finding small instances~\cite{Fast_Tiny}. In DOTA-v2.0, the sizes of newly collected GF-2 satellite images and CycloMedia airborne images are usually $29,200\times27,620$ and $7,360\times4,912$ pixels, respectively.

			\subsection{Category Selection}
			We choose eighteen categories, {\em plane, ship, storage tank, baseball diamond, tennis court, swimming pool, ground track field, harbor, bridge, large vehicle, small vehicle, helicopter, roundabout, soccer ball field, basketball court, container crane, airport and helipad.}
			We select these categories according to their frequency of occurrence and value for real-world applications. The first 10 categories are common in the existing datasets, \eg,~\cite{DLR3KMunichVehicle,VHR,ucas-aod,COWC}. 
			Other categories are added considering their value in real-world applications. For example, we selected ``helicopter'' as moving objects are of significant importance in aerial images, and ``roundabout'' as it plays an essential role in roadway analyses.
			It is worth discussing whether to take ``stuff" categories into account. There are usually no clear definitions for the "stuff" categories (\eg {\em harbor, airport, parking lot}), as shown in the Scene UNderstanding (SUN) dataset~\cite{SUN}. However, their contextual information may be helpful for object detection.
			Based on this idea, we select the harbor and airport categories because their borders are relatively easy to define and there are abundant harbor and airport instances in our image sources.
			
			\subsection{Oriented Object Annotation}\label{sec:annotation_method}
				\begin{figure}[t!]
				\centering
				\includegraphics[width=0.82\linewidth]{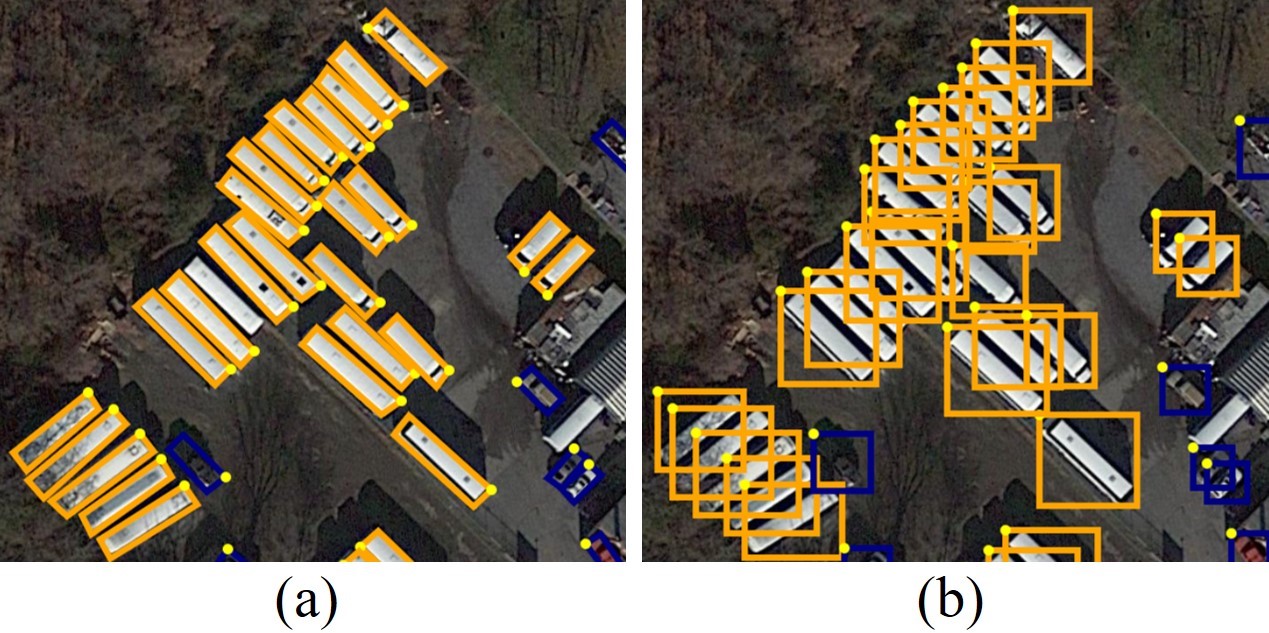}
				\vspace{-4mm}
				\caption{{\bf Comparisons between HBB and OBB representations for objects.} (a) OBB representation. (b) HBB representation. The HBB representation cannot distinguish rotated dense objects. 
				}
				\label{fig:compare_annotations}
				\vspace{-3mm}
			\end{figure}

			In computer vision, many visual concepts, such as region descriptions, objects, attributes, and relationships, are often represented with bounding boxes, as shown in~\cite{boundingboxannotation}. A common representation of the bounding box is \((x_c,y_c,w,h)\), where \((x_c, y_c)\) is the center location and \(w,h\) are the width and height, respectively, of the bounding box. We call this type of bounding box an HBB. The HBB can describe objects well in most cases. However, it cannot accurately outline oriented instances such as text and objects in aerial images. 
			As shown in Fig.~\ref{fig:compare_annotations}, the HBB cannot differentiate densely-distributed oriented objects. The conventional NMS algorithm fails in such cases. On the other hand, the regional features extracted from HBBs are not rotation invariant. 
			To address these problems, we represent the objects with OBBs. In detail, an OBB is denoted by \(\left \{ (x_i,y_i)| i=1,2,3,4 \right \}\), where \((x_i, y_i)\) denotes the position of the OBB's vertex in the image. The vertices are arranged in clockwise order. 
				
				The most straightforward way to annotate an OBB is to draw an HBB and then adjust the angle. However, since there is no reference for HBBs, several adjustments in the center, height, width and angle are usually needed to fit an arbitrarily oriented object well. Clicking on physical points lying on the object~\cite{extremeclicking} could make crowd-sourced annotations more efficient for HBBs, as these points are easy to find. Inspired by this idea, we allow the annotators to click four corners of the OBBs. For most categories, the corners of the OBBs (\eg, tennis court and basketball court) lie on or close to the objects (vehicles), however, there are still some categories whose shapes are very different from OBBs. For these categories, we annotate four key points lying on the object. For example, we annotate the planes with 4 key points, representing the head, two wingtips, and tail. Then we transfer the 4 key points to an OBB.

				However, when using OBBs to represent objects, we could obtain four different representations for the same object by changing the order of the points. For example, assume that \( (x_1, y_1, x_2, y_2, x_3, y_3, x_4, y_4) \) represents an object, but we could represent the same object by \((x_2, y_2, x_3, y_3, x_4, y_4, x_1, y_1)\). For categories having differences between the head and tail (\eg, {\it helicopter, large vehicle, small vehicle, harbor}), we carefully select the first point to imply the ``head'' of the object.
				For other categories (\eg, {\it soccer-ball field, swimming pool and bridge}) that do not have visual clues to determine the first point, we choose the top-left point as the starting point.
				
\begin{figure*}[t!]
					\centering					\includegraphics[width=0.88\linewidth]{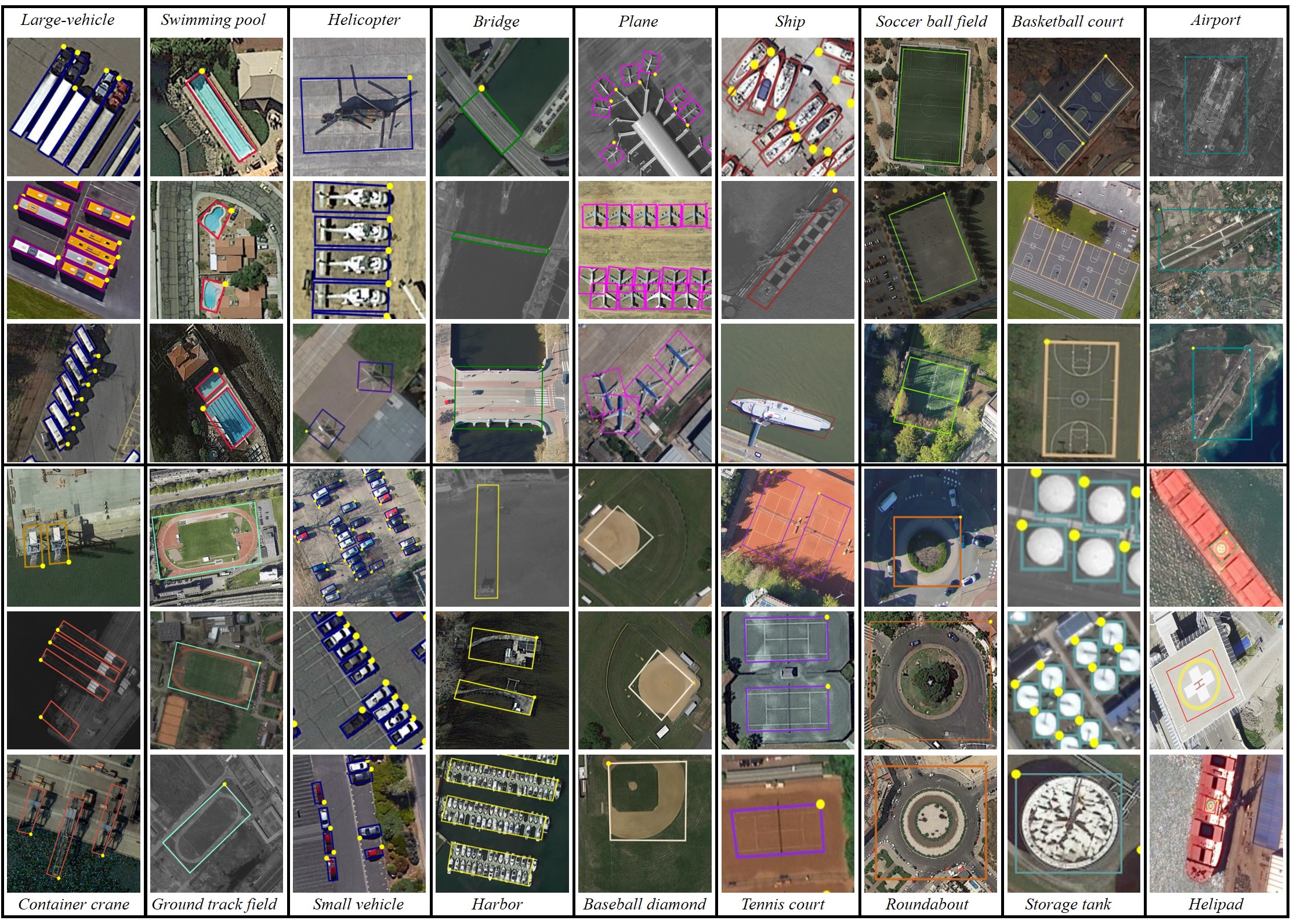}					\vspace{-3mm}
					\caption{Examples of annotated images in DOTA. We show three examples per category. }
					\label{fig:samples}
					\vspace{-4mm}
				\end{figure*}

				\jian{The detailed pipeline is described in the following. First, we developed a customized annotation tool. Based on this, we asked experts in the interpretation of aerial images to annotate some examples for each category. The annotations by the experts are used to train the volunteers for the large-scale annotation. 
				After the training, we evaluate the annotation ability of volunteers to separate them into the \emph{plain} and \emph{senior} groups. The volunteers in the plain group are asked to yield the initial annotations which are doubly-checked by senior volunteers and the authors. The images that do not pass the checking were sent back to volunteers to improve the annotation quality. The volunteers were mainly recruited from Wuhan University, with a background in remote sensing image interpretation.} Some examples of annotated patches are shown in Fig.~\ref{fig:samples}.
				
				\jian{\textbf{Discussion.} There are two types of possible errors in the object annotations: 1) missing annotations; 2) inaccurate bounding boxes annotations. Missing annotations is mainly caused by the difficulty in identifying tiny objects. The proportion of missed objects can be ignored and does not influence the training and evaluation of object detectors in DOTA-v2.0. However, if the researchers want to study this problem and further improve the performance, we recommend researchers refer to these prior works~\cite{softsampling,solvingmissing}. The inaccurate bounding boxes annotations exist in all object detection datasets since there exists ambiguity to define the boundary of objects sometimes (\eg,~the occlusion). Developing algorithms that modeling the inaccurate bounding boxes annotation has been studied in~\cite{softernms} for natural images. However, He {\em et al}.~\cite{softernms} only studied the inaccurate HBB annotation. The modeling of inaccurate annotation of the OBBs in DOTA can be researched in the future.}
				
				\section{Properties of DOTA}
				\label{statistics_DOTA}
				
				\subsection{Image Sources}

\begin{table}[t!]
\caption{The statistics for the annotated objects across different data sources in DOTA-v1.5 and DOTA-v2.0. The total image areas, the area of objects and the ratio of foreground pixels of the annotated objects to the image areas are reported.}
\label{tab:foregorund}
\centering
\vspace{-3mm}
\begin{tabular}{ccccc}
\hline
\multicolumn{5}{c}{DOTA-v1.5}                                                                                                                        \\ \hline
\multicolumn{1}{c|}{}                      & \multicolumn{1}{c|}{Google Earth} & \multicolumn{1}{c|}{GF$\&$JL}   & \multicolumn{1}{c|}{Aerial} & All     \\ \hline
\multicolumn{1}{c|}{\# of images}          & \multicolumn{1}{c|}{2375}         & \multicolumn{1}{c|}{431}    & \multicolumn{1}{c|}{/}      & 2806    \\
\multicolumn{1}{c|}{Images Area ($10^6$)}  & \multicolumn{1}{c|}{11,873}       & \multicolumn{1}{c|}{7,301}  & \multicolumn{1}{c|}{/}      & 19,173  \\
\multicolumn{1}{c|}{Objects Area ($10^6$)} & \multicolumn{1}{c|}{784}          & \multicolumn{1}{c|}{20}     & \multicolumn{1}{c|}{/}      & 804     \\
\multicolumn{1}{c|}{Foreground Ratio}           & \multicolumn{1}{c|}{0.066}        & \multicolumn{1}{c|}{0.003}  & \multicolumn{1}{c|}{/}      & 0.042   \\ \hline
\multicolumn{5}{c}{DOTA-v2.0}                                                                                                                        \\ \hline
\multicolumn{1}{c|}{\# of images}          & \multicolumn{1}{c|}{10186}        & \multicolumn{1}{c|}{516}    & \multicolumn{1}{c|}{566}    & 11268   \\
\multicolumn{1}{c|}{Images Area ($10^6$)}  & \multicolumn{1}{c|}{29,991}       & \multicolumn{1}{c|}{75,854} & \multicolumn{1}{c|}{20,462} & 126,306 \\
\multicolumn{1}{c|}{Objects Area ($10^6$)} & \multicolumn{1}{c|}{1,111}        & \multicolumn{1}{c|}{243}    & \multicolumn{1}{c|}{673}    & 2,027   \\
\multicolumn{1}{c|}{Foreground Ratio}           & \multicolumn{1}{c|}{0.037}        & \multicolumn{1}{c|}{0.003}  & \multicolumn{1}{c|}{0.033}  & 0.016   \\ \hline
\end{tabular}
\vspace{-4mm}
\end{table}
				\jian{The images in DOTA-v2.0 are from three different sources, \ie, Google Earth images, GF-2 and JL-1 (GF$\&$JL) satellite images, and the CycloMedia~\cite{cyclomedia} airborne images. Tab.~\ref{tab:foregorund} shows the statistics of three image sources in terms of the images area, objects area, and foreground ratio. 
				}
				We can see that the carefully selected Google Earth images contain the majority of positive samples. Nevertheless, the negative samples are also important to avoid positive sample bias~\cite{torralba2011unbiased}. The object distributions in the collected GF\&JL satellite images and CycloMedia airborne images are close to those in real-world applications and provide enough background area.
				\jian{It is worthwhile to notice that DOTA-v2.0 contains both RGB images and grayscale images. More precisely, the images collected from Google Earth and CycloMedia are often RGB-rendered versions of original aerial images, and the images from GF-2, JL-1 are 8-bit per pixel optimally converted from their original panchromatic band in 10-bit. 
				However, during those spectral rendering and bit-length optimization processes, the structure and appearance information of the image content are always consistent and the images are feasible for recognition-oriented tasks~\cite{AID}.}

				\jian{The acquisition dates are available for all the images from GF-2, JL-1, and CycloMedia, and for $27\%$ of the images collected from Google Earth. As the main goal of our task is to recognize objects in aerial images by relying on visual cues, for which the geolocation of an image is insignificant for the process, DOTA-v2.0 does not provide the geolocation of its images.}
				
			\begin{figure}[t!]
				\centering				\includegraphics[width=0.9\linewidth]{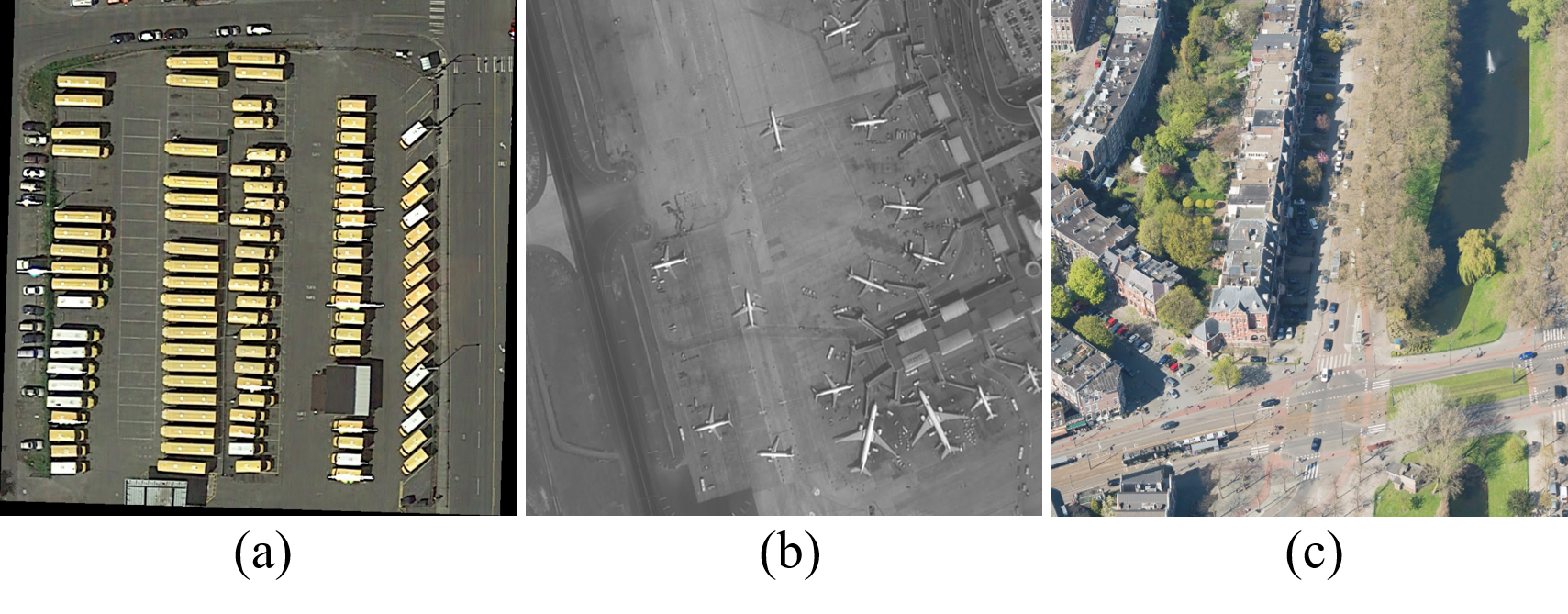}
				 \vspace{-4mm}
				\caption{Typical examples of images collected from Google Earth (a), GF\&JL satellite (b) and CycloMedia (c).}
				\label{fig:image_sources}
				\vspace{-3mm}
			\end{figure}
				
				
				\vspace{-3mm}
				\subsection{Spatial Resolution Information}
\begin{figure}[t!]
            		\centering
	\includegraphics[height = 0.4 \linewidth,width=0.8\linewidth]{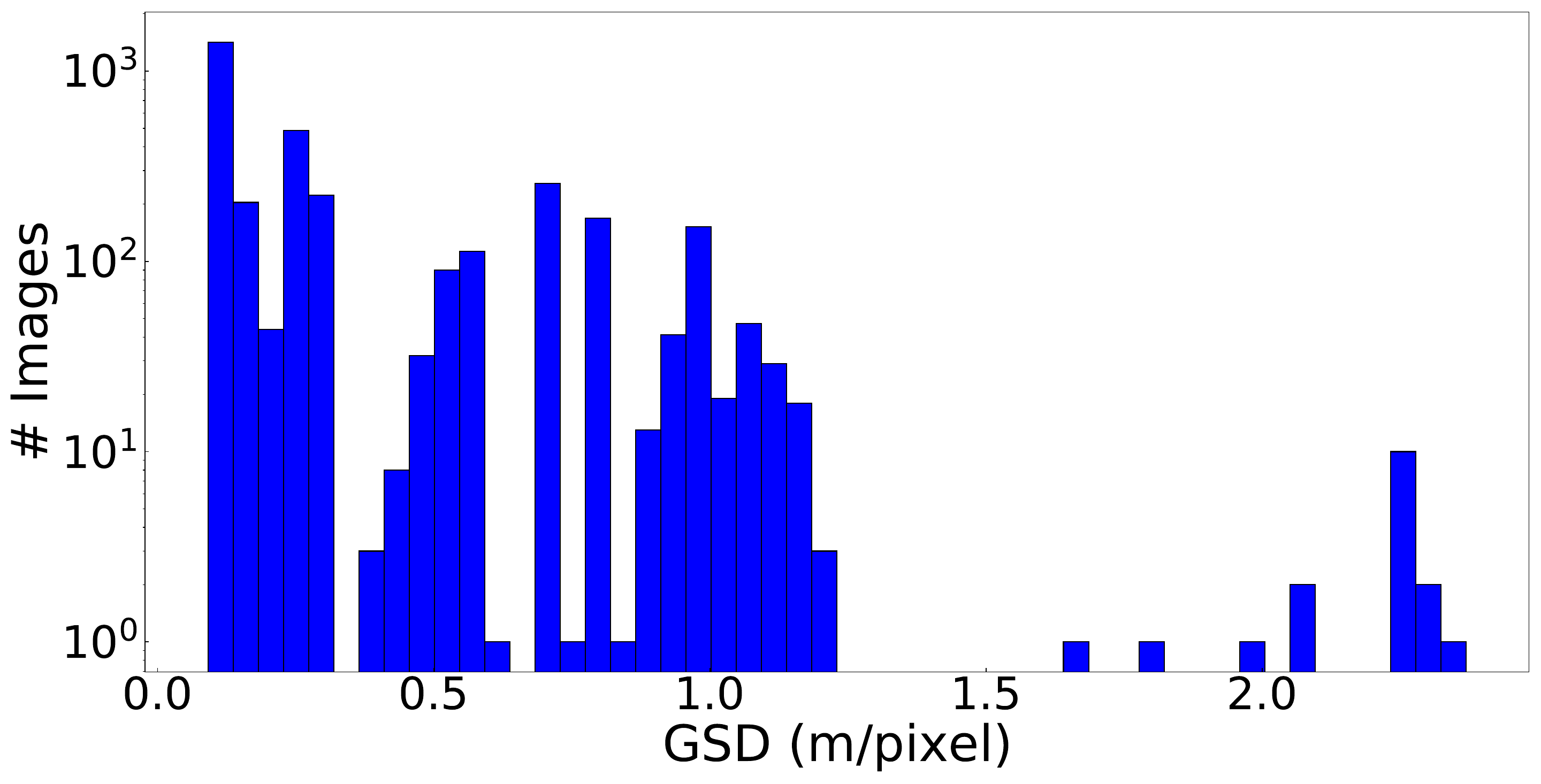}
            		    \vspace{-3mm}       			\caption{The statistics of the GSD \DingJian{in $30\%$ of the images in DOTA-v2.0.} }
            		    \label{fig:gsd_distribution}
            			\vspace{-4mm}
            	\end{figure}

				The ground sample distance (GSD), which indicates the distance between pixel centers measured on Earth, has potential usages.
	              For example, it allows us to calculate the actual sizes of objects, which can be used to filter mislabeled or misclassified outliers since the object sizes of the same category are usually limited to a small range. \DingJian{The GSD can also be directly incorporated into object detectors~\cite{gsdet} to improve the classification accuracy of categories that have less physical size variation.} Furthermore, we can conduct scale normalization~\cite{snip} based on the priors of the object size and GSD. 
	              \jian{In DOTA-v2.0, the GSDs of the images from GF-2, JL-1, and CycloMedia are 0.81, 0.72, and 0.1 meters per pixel, respectively. While the GSDs of the images from Google Earth range from 0.1m to 4.5m per pixel. The statistical distribution of GSDs is shown in Fig.~\ref{fig:gsd_distribution}. It is noted that only $30\%$ of the images in DOTA-v2.0 have the GSD information.} \DingJian{However, the missing of GSDs will not have a big impact on the applications that require GSDs, since a learning-based method can be used to estimate the GSD~\cite{gsd}. }
	              
				\subsection{Various Instance Orientations}
			    Objects in the overhead view images have a high diversity of orientations without the restriction of gravity. As shown in Fig.~\ref{fig:large-example}~(g), the objects have equal probabilities of arbitrary angles in $[-\pi, \pi]$. It is worthwhile to note that although objects in scene text detection~\cite{icdar15} and face detection~\cite{rotation_face} also have many orientation variations, the angles of most objects lie within a narrow range (\eg, $[-\pi/2, \pi/2]$) due to gravity. The unique angle distributions of DOTA make it a good dataset for research on rotation-invariant feature extraction and oriented object detection.
				\subsection{Various Instances Pixel Sizes}
				Following the convention in~\cite{WIDERFACE}, we use the height of an HBB to measure the pixel size of the instance. We divide all the instances in our dataset into three splits according to their heights of HBBs: small, with range from $10$ to $50$, medium, with range from $50$ to $300$, and large, with range above $300$.
				Tab.~\ref{fig:size-distribution} illustrates the percentages of these three instance splits in different datasets. It is clear that the PASCAL VOC dataset,  NWPU VHR-10 dataset and DLR 3K Munich Vehicle dataset are dominated by medium instances or small instances. 

				MS COCO and DOTA-v1.0 have a good balance between small instances and medium instances.
				DOTA-v2.0 has more small instances than DOTA-v1.0. In DOTA-v2.0, some instances that are approximately 10 pixels are annotated.

\begin{figure}[t!]
	\centering					\includegraphics[width=0.92\linewidth]{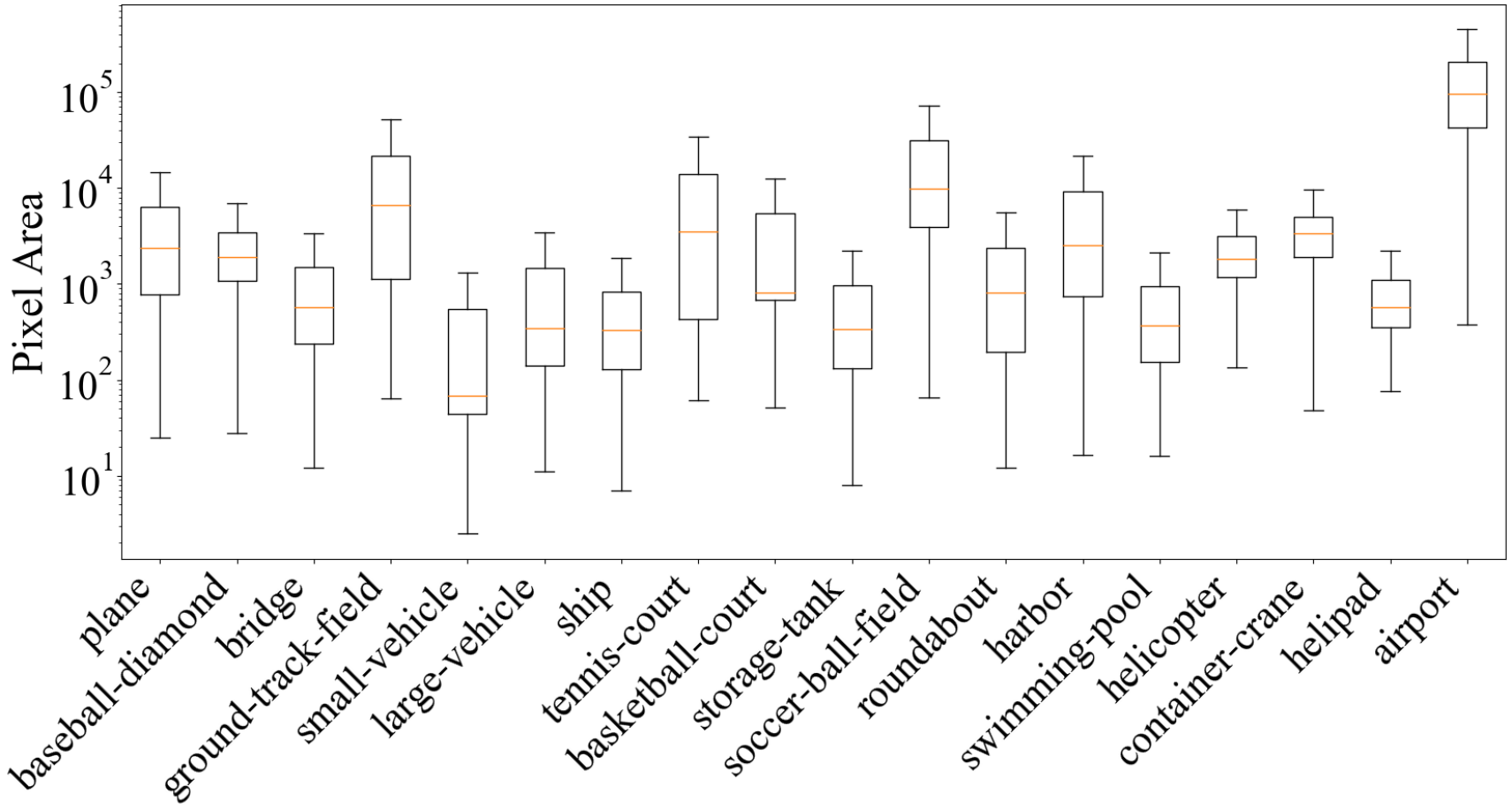}
	\vspace{-3mm}
	\caption{Size variations for each category in DOTA. The sizes of different categories vary in different ranges. }
	\label{fig:boxplot}
	\vspace{-3mm}
	\end{figure}

				In Fig.~\ref{fig:boxplot}, we also show the distribution of instances' pixel sizes for different categories in DOTA. This figure indicates that the scales vary greatly both within and between categories. These large-scale variations among instances make the detection task more challenging.
				\begin{table}[t!]
					\footnotesize
					\centering
					\caption{Comparison of the instance size distributions of aerial and natural images in some datasets.}
					\vspace{-3mm}
\begin{tabular}{cccc}
\hline
Dataset                          & 10-50 pixels   & 50-300 pixels  & $>$300 pixels \\ \hline
PASCAL VOC~\cite{PASCALVOC}      & 0.14          & 0.61          & 0.25            \\
MS COCO~\cite{COCO}               & 0.43          & 0.49          & 0.08            \\ \hline
NWPU VHR-10~\cite{VHR}           & 0.15          & 0.83          & 0.02            \\
DLR 3K~\cite{DLR3KMunichVehicle} & 0.93          & 0.07          & 0               \\
DOTA-v1.0~\cite{DOTA}            & 0.57          & 0.41          & 0.02            \\
DOTA-v1.5                        & 0.79          & 0.2           & 0.01            \\
DOTA-v2.0                        & 0.77 & 0.22 & 0.01   \\ \hline
\end{tabular}
\label{fig:size-distribution}
\vspace{-3mm}
\end{table}
				
				\subsection{Various Instance Aspect Ratios (ARs)}
\begin{figure}[t!]
				\centering					\includegraphics[width=0.92\linewidth]{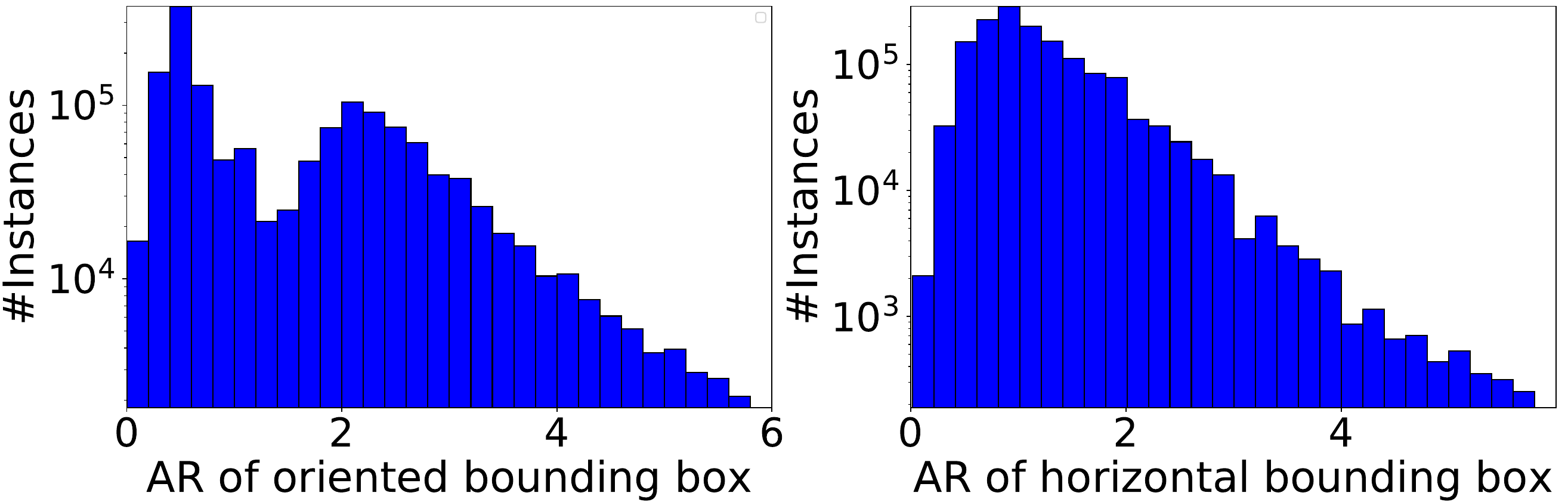}
				    \vspace{-3mm}
					\caption{AR distributions of the instances in DOTA. (a) The ARs of the OBBs. (b) The ARs of the HBBs.}
					\label{fig:ar_instances}
					\vspace{-3mm}
				\end{figure}

				The AR is essential for anchor-based models, such as Faster R-CNN~\cite{FasterR-CNN} and You Only Look Once (YOLOv2)~\cite{yolo9000}. 
				We use two kinds of ARs for all the instances in our dataset to guide the model design namely, 1) the ARs of the original OBBs and 2) the AR of HBBs, which are generated by calculating the axis-aligned bounding boxes over the OBBs.  
				Fig.~\ref{fig:ar_instances} illustrates the distributions of these two types of aspect ratios in DOTA. We can see that instances vary significantly in aspect ratio. Moreover, many instances have a large aspect ratio in our dataset.

				\begin{figure}[t!]
					\centering
					\includegraphics[width=0.92\linewidth]{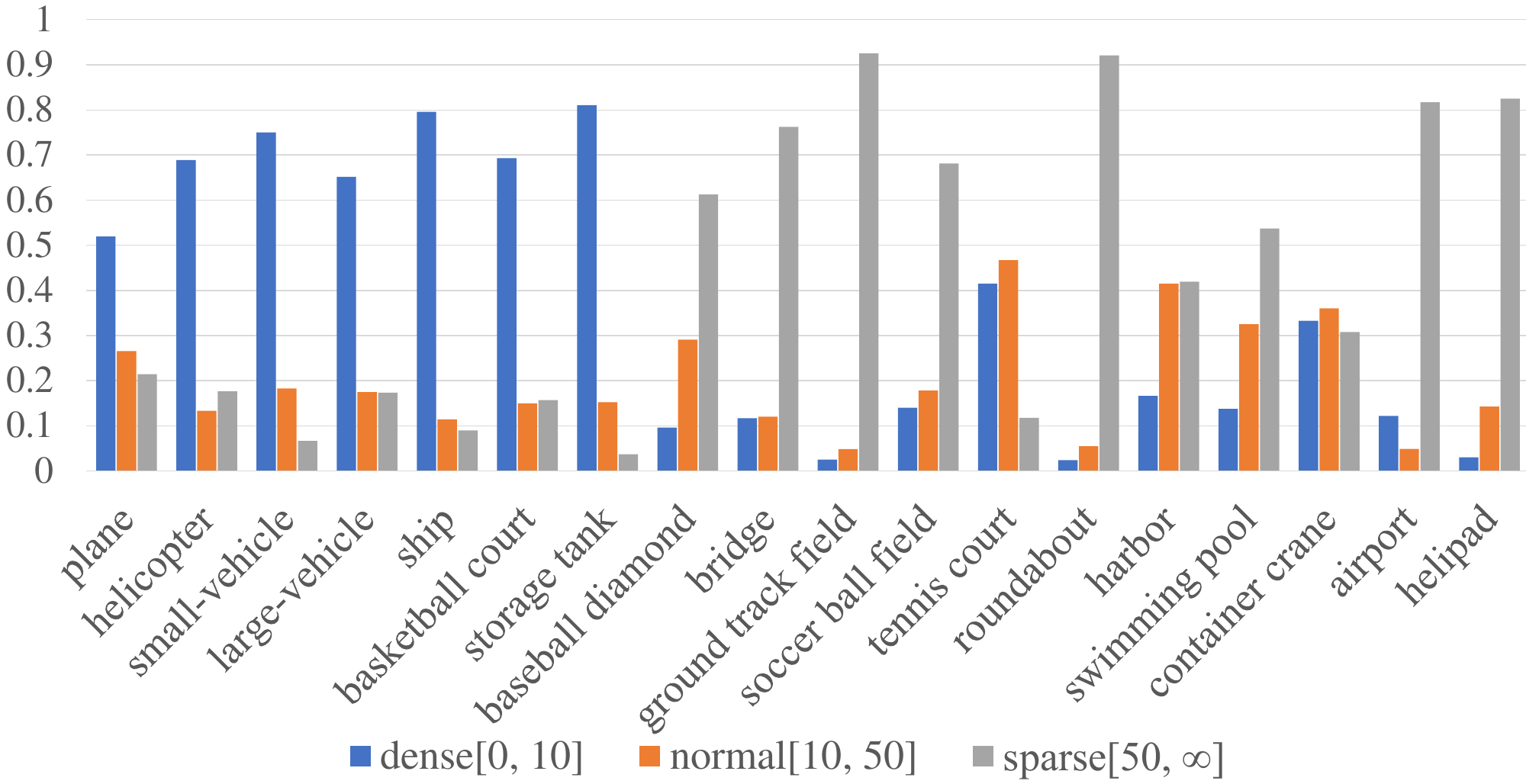}
					\vspace{-3mm}
					\caption{Densities of the different categories. The density is measured by calculating the distance to the closest instance.}
					\label{fig:dense_dist}
					\vspace{-3mm}
				\end{figure}
				
				\subsection{Various Instance Densities of the Images}

				The number of instances per image is an important property for object detection datasets and varies largely in DOTA. It can be very dense (up to 1000 instances per image patch), or very sparse (only one instance per image patch). 
We compare this property among DOTA and the general object detection datasets in Fig.~\ref{compare_general_dataset}. 
The number of instances per image in DOTA varies more widely than in natural image datasets.

Different categories also have different density distributions. To give a quantitative analysis, for each instance, we first measure the distance to the closest instance in the same category. We then bin the distances into three parts, dense $[0, 10)$, normal $[10, 50)$ and sparse $[50, \infty)$ (see Fig.~\ref{fig:dense_dist}). Fig.~\ref{fig:dense_dist} shows that the storage tank, ship and small vehicle are top-3 dense categories.
				
				\subsection{DOTA Versions}
				\jian{It is important to notice the significant improvements from DOTA-v1.0 to DOTA-v2.0. In DOTA-v1.0, tiny objects (below 10 pixels) have not been annotated, and images are mainly from a single domain, \ie, Google Earth images. Moreover, the images from DOTA-v1.0 are usually selected \textit{areas that contain many objects} from large-size images.
		        Although, in the past years, promising progress has been reported in oriented object detection with DOTA-v1.0, following challenging aspects can not been fully addressed by using DOTA-v1.0:
		        \begin{itemize}
		            \item[-] to benchmark detection models for oriented objects both in tiny size and normal size;
		            \item[-] to address the object detection problem in large-scale images, \eg, images with size larger than $20,000\times 20,000$ pixels,  that only contain a few objects; 
		            \item[-] to develop robust oriented object detection models with strong generalization capability for multi-source overhead images.
		        \end{itemize}
		        To address these problems, DOTA-v1.5 added the annotation of the tiny objects in DOTA-v1.0. DOTA-v2.0 further collected many more large-size GF-2 and airborne images, which have a lower foreground ratio, approaching the object distribution in real-world applications, as shown in Tab.~\ref{tab:foregorund}. The number of objects for each category and the dataset split for three versions of DOTA are summarized in Tab.~\ref{tab:compare_dota_versions}.}
		        

\begin{table}[t!]
\caption{Comparisons of the three versions of DOTA. We count the number of instances for each category and dataset split.}
\vspace{-3mm}
\centering
\begin{tabular}{c|c|c|c}
\hline
               & DOTA-v1.0 & DOTA-v1.5 & DOTA-v2.0 \\ \hline
Plane          & 14,085     & 14,978     & 23,930     \\
BD             & 1,130      & 1,127      & 3,834      \\
Bridge         & 3,760      & 3,804      & 21,433     \\
GTF            & 678       & 689       & 4,933      \\
SV             & 48,891     & 242,276    & 1,235,658    \\
LV             & 31,613     & 39,249     & 89,353     \\
Ship           & 52,516     & 62,258     & 251,883    \\
TC             & 4,654      & 4,716      & 9,396      \\
BC             & 954       & 988       & 3,556     \\
ST             & 11,794     & 12,249     & 79,497     \\
SBF            & 720       & 727       & 2,404      \\
RA             & 871       & 929       & 6,809      \\
Harbor         & 12,287     & 12,377     & 29,581     \\
SP             & 3,507      & 4,652      & 20,095    \\
HC             & 822       & 833       & 893       \\
CC             & 0         & 237       & 3,887      \\
Airport        & 0         & 0         & 5,905      \\
Helipad        & 0         & 0         & 611       \\
Total          & 188,282    & 402,089    & 1,793,658   \\ \hline
Training          & 98,990     & 210,631    & 268,627    \\
Validation            & 28,853     & 69,565     & 81,048     \\
Test/Test-dev  & 60,439     & 121,893    & 353,346    \\
Test-challenge & 0         & 0         & 1,090,637    \\ \hline
\end{tabular} \label{tab:compare_dota_versions}
\vspace{-2mm}
\end{table}

            \subsubsection{DOTA-v1.0}
				DOTA-v1.0 contains 15 common categories, 2,806 images and 188, 282 instances. The proportions of the training set, validation set, and testing set in DOTA-v1.0 are 1/2, 1/6, and 1/3, respectively.
				
				\subsubsection{DOTA-v1.5}
				DOTA-v1.5 uses the same images as DOTA-v1.0, but extremely small instances (less than 10 pixels) are also annotated. Moreover, a new category, ``container crane" containing 402,089 instances in total is added. The number of images and dataset splits are the same as those in DOTA-v1.0. 
				
				\subsubsection{DOTA-v2.0}
				There are 18 common categories, 11,268 images and 1,793,658 instances in DOTA-v2.0. Compared to DOTA-v1.5, it further adds the new categories of ``airport" and ``helipad".
				DOTA-v2.0 are split into {\em training, validation, test-dev}, and {\em test-challenge} subsets. To avoid the problem of over-fitting, the proportion of the training and validation sets is smaller than that of the test set. Furthermore, we have two test subsets, namely {\em test-dev} and {\em test-challenge}, which are similar to the MS COCO dataset~\cite{COCO}. 
				
				For the test-dev subset and the test-challenge subset, we release the images without annotations. For evaluation, one can submit the results to the evaluation server\footnote{\url{https://captain-whu.github.io/DOTA/evaluation.html}}. All the DOTA-v2.0 experiments in this paper are evaluated on test-dev. 
				
				\section{Benchmarks}
				\subsection{Evaluation Tasks and Metrics}\label{evaluation_metric}
				
			\begin{figure}[t!]
				\centering				\includegraphics[width=0.6\linewidth]{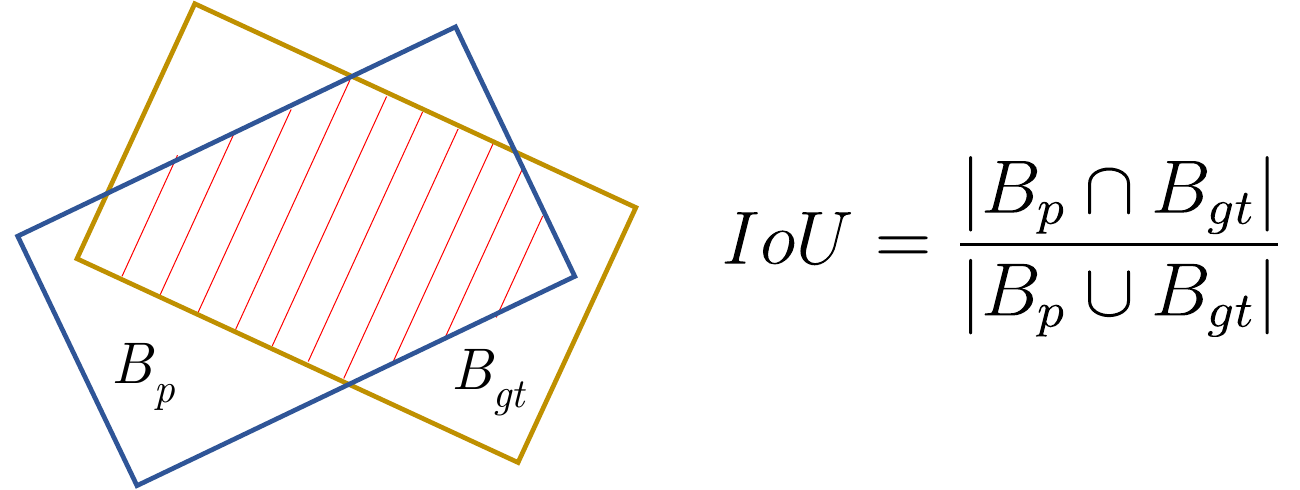}
				\caption{The computation of the IoU between two OBBs.}
				\label{fig:polygonIoU}
				\vspace{-3mm}
			\end{figure}
			
 The task of object detection is to locate and classify the instances in images. 
 We use two location representations (\textbf{HBB} and \textbf{OBB}) in our paper. The HBB is a rectangle \((x, y, w, h)\), \jian{and the OBB is a quadrilateral \(\{(x_i, y_i)|i=1,2,3,4\}\).}
Then, there are two tasks, detection with HBB and detection with OBB. To be more specific, we evaluate these methods on two kinds of ground truths: HBB and OBB ground truths. \jian{For the two tasks, each detected bounding box has a corresponding confidence score.} We adopt the PASCAL VOC 07 metric~\cite{PASCALVOC} for the calculation of the mean average precision (mAP). 
			\jian{Average Precision (AP) computes the average precision value for recall value over 0 to 1 (\ie, the area under precision/recall curve). The mean Average Precision (mAP) is the average of AP over all classes. The detailed computation of the precision and recall can refer to~\cite{PASCALVOC}. The intersection over union (IoU) is crucial in determining true positives and false positives, which are required to compute precision and recall.} It is worthwhile to note that for the OBB task, the intersection over union (IoU) is calculated between OBBs\jian{, as shown in Fig.~\ref{fig:polygonIoU}. The two OBBs ($B_p$ and $B_{gt}$), and the intersection between OBBs ($B_p\bigcap B_{gt}$) are all convex polygons, whose area can be easily computed\footnote{A polygon can be decomposed into a group of non-overlapping triangles. The area of the convex polygon is equal to the sum of all the triangular areas.}. The union area of two OBBs can be calculated as $|B_p \bigcup B_{gt}|= |B_p| + |B_{gt}| - |B_p \bigcap B_{gt}|$. The code for the mAP and IoU computation between OBBs can be found in our development kit.}
				
				\subsection{Implementation Details} \label{sec:benchmarks}
				In the previous benchmarks~\cite{DOTA}, the algorithms were implemented with different codes and settings, which makes these algorithms hard to compare in DOTA.
				To this end, we implement and evaluate all the algorithms in one unified code library modified from MMDetection~\cite{mmdetection}.

				Since large images cannot be directly fed to CNN-based detectors due to the memory limitations, we crop a series of \(1,024\times1,024\) patches from the original images with a stride set to 824 (different from the previous stride of 512~\cite{DOTA}). 
				During inference, we first send the patches (same settings as training) to obtain temporary results. Then we map the detected results from the patch coordinates to the original image coordinates. Finally, we apply NMS on these results in the original image coordinates. We set the threshold of NMS to $0.3$ for the HBB experiments and $0.1$ for the OBB experiments.
				For multi-scale training and testing, we first scale the original images to $[0.5, 1.0, 1.5]$ and then crop the images into patches of size $1,024 \times 1,024$ and a stride of 824.
				We use 4 GPUs for training with a total batch size of 8 (2 images per GPU). The learning rate is set to 0.01. Except for RetinaNet~\cite{focal}, which adopts the "$2\times$" schedule, the other algorithms adopt the "$1\times$"~\cite{Detectron2018} training schedule. We set the number of proposals and maximum number of predictions per image patch to 2,000 for all the experiments except when otherwise mentioned.
				The other hyperparameters follow those of Detecron~\cite{Detectron2018}.

				\subsubsection{Baselines with HBBs}
				\label{sec:horizontal_baseline}

We use two ways to build baselines for the HBB task. The first way directly predicts the HBB results, while the second way first predicts the OBB results and then converts OBBs to HBBs. To directly predict the HBB results, we use RetinaNet~\cite{focal}, Mask R-CNN, Cascade Mask R-CNN, Hybrid Task Cascade and Faster R-CNN~\cite{FasterR-CNN} as baselines.
For the OBB predictions, we will introduce the methods in the following section.
				\subsubsection{Baselines with OBBs}
				\label{obb_baseline_frame}
				Most of the state-of-the-art object detection methods are not designed for oriented objects. To enable these methods to predict OBBs, we build the baselines in two ways. The first is to change HBB head to OBB Head, which regresses the offsets of OBBs relative to the HBBs. The second is Mask Head, which considers the OBBs to a coarse mask and predicts the pixel-level classification from each RoI. 
			
\textbf{OBB Head} To predict OBB, the previous Faster R-CNN OBB~\cite{DOTA} and Textboxes++~\cite{textboxes++} modified RoI Head of Faster R-CNN and the Anchor Head of the single-shot detector (SSD), respectively, to regress quadrangles. In this paper, we use the representation \((x, y, w, h, \theta)\) instead of \(\{(x_i, y_i)|i=1,2,3,4\}\) for OBB regression. More precisely, rectangular RoIs (anchors) can be written as \(\mathcal{B} = (x_{min}, y_{min}, x_{max}, y_{max})\). We can also consider it a special OBB and rewrite it as \(\mathcal{R} = (x, y,\)\( w, h, \theta)\). 
                For matching, IoUs are calculated between the horizontal RoIs (anchors) and HBBs of the ground truths for computational simplicity.
                Each OBB, it has four forms: \(\mathcal{G} = \{gt_i|i=1,2,3,4\}\), where \(gt_1 = (x_g, y_g, w_g, h_g, \theta_g)\), \(gt_2 = (x_g, y_g, w_g, h_g, \theta_g + \pi)\), \(gt_3 = (x_g, y_g, h_g, w_g, \theta_g)\), and \(gt_4 = (x_g, y_g, h_g, w_g, \theta_g + \pi)\). Before calculating the targets, we choose the best matched ground-truth form. The index of the best matched form is calculated by $\xi = \mathop{\arg\min\limits_{i}}  \mathcal{D} (\mathcal{R}, gt_{i})$, where $\mathcal{D}$ is a distance function, which could be Euclidean distance or another distance function. We denote the best matched form by \(gt_\xi = (x_b, y_b, w_b, h_b, \theta_b)\).  Then the learning target 
                is calculated as
                \begin{align}
                t_{x} &= (x_{b} - x) /  w, \; t_{y} = (y_{b} - y) / h, \nonumber\\
                t_{w} &= log(w_b/w), \; t_{h} = log(h_b/h), \\
                t_{\theta} &= \theta_b - \theta \nonumber
                \end{align}
                We then simply replace the HBB RoI Head of \textit{Faster R-CNN} and anchor head of \textit{RetinaNet} with OBB Head and obtain two models, called \textit{Faster R-CNN OBB} and \textit{RetinaNet OBB}. We also modify the Faster R-CNN to predict both the HBB and OBB in parallel, which is similar to Mask R-CNN~\cite{MaskR-CNN}. We call this model\textit{ Faster R-CNN H-OBB}. We further evaluate the deformable RoI pooling (Dpool) and RoI Transformer by replacing the RoI Align in Faster R-CNN OBB. Then we have two models: \textit{Faster R-CNN OBB + Dpool} and \textit{Faster R-CNN OBB + RoI Transformer}. Note that the RoI Transformer used here is slightly different from the original one. The original RoI Transformer uses the Light Head R-CNN~\cite{light_head} as the base detector while we use Faster R-CNN.
				
				\textbf{Mask Head} Mask R-CNN~\cite{MaskR-CNN} was originally used for instance segmentation. Although DOTA does not have pixel-level annotation for each instance, the OBB annotations can be considered coarse pixel-level annotations, so we can apply Mask R-CNN~\cite{MaskR-CNN} to DOTA. During inference, we calculate the minimum OBBs that contain the predicted masks.
				The original Mask R-CNN~\cite{MaskR-CNN} only applies a mask head to the top 100 HBBs in terms of the score. Due to the large number of instances per image, as illustrated in Fig.~\ref{fig:instance_per_image_compare}, we apply a mask head to all the HBBs after NMS. In this way, we evaluate Mask R-CNN~\cite{MaskR-CNN}, Cascade Mask R-CNN and Hybrid Task Cascade~\cite{hybridcascade}.
				
				\begin{table*}[t!]
					\centering
					\caption{Baseline results on DOTA. For the evaluation of DOTA-v2.0, we use the DOTA-v2.0 test-dev set. The implementation details are described in Sec.~\ref{sec:benchmarks}. All the algorithms in this table adopt the ResNet-50 with an FPN as backbone. The speed refers to the inference speed, which is reported for a single NVIDIA Tesla V100 GPU on DOTA-v2.0 test-dev. The image size is $1,024\times 1,024$. Hybrid Task Cascade* means that the semantic branch is not used since there are no semantic annotations in DOTA.}
					\vspace{-3mm}
					\resizebox{0.95\linewidth}{!}{
					\begin{tabular}{c|c|cc|cc|cc}
						\hline \hline
						\multirow{2}{*}{method}          & \multirow{2}{*}{speed (fps)} & \multicolumn{2}{c|}{DOTA-v1.0} & \multicolumn{2}{c|}{DOTA-v1.5} & \multicolumn{2}{c}{DOTA-v2.0} \\ \cline{3-8} 
						&                        & HBB mAP        & OBB mAP       & HBB mAP        & OBB mAP       & HBB mAP        & OBB mAP       \\ \hline
						RetinaNet~\cite{focal}                        & 16.7                & 67.45          & -          & 61.64          & -          & 49.31          & -          \\
						RetinaNet OBB~\cite{focal}                    & 12.1               & 69.05          & 66.28         & 62.49          & 59.16         & 49.26          & 46.68         \\
					Mask R-CNN~\cite{MaskR-CNN}                        & 9.7                 & 71.61          & 70.71         & 64.54          & 62.67         & 51.16          & 49.47         \\
						Cascade Mask R-CNN~\cite{hybridcascade}                & 7.2                 & 71.36          & 70.96         & 64.31          & 63.41         & 50.98          & 50.04         \\
						Hybrid Task Cascade*~\cite{hybridcascade}             & 7.9                 & 72.49          & 71.21         & 64.47          & 63.40         & 50.88          & 50.34         \\
						Faster R-CNN~\cite{FasterR-CNN}                      & 14.3                & 70.76          & -          & 64.16          & -          & 50.71          & -          \\
						Faster R-CNN OBB~\cite{DOTA}                  & 14.1                & 71.91          & 69.36         & 63.85          & 62.00         & 49.37          & 47.31         \\
						Faster R-CNN OBB + Dpool~\cite{Deformable}          & 12.1                & 71.83          & 70.14         & 63.67          & 62.20         & 50.48          & 48.77         \\
						Faster R-CNN H-OBB~\cite{DOTA}                & 13.7                & 70.37          & 70.11         & 64.43          & 62.57         & 50.38          & 48.90         \\
						Faster R-CNN OBB + RoI Transformer~\cite{RoITransformer} & 12.4                  & \textbf{74.59}        & \textbf{73.76}       & \textbf{66.09}         & \textbf{65.03}       & \textbf{53.37}       & \textbf{52.81}        \\ \hline
					\end{tabular} 
					}
					\label{benchmarks}
					\vspace{-3mm}
				\end{table*}
				
			\begin{table*}[t!]
				\footnotesize
				\caption{Baseline results of class-wise AP on DOTA-v1.0. The abbreviations of algorithms are: Mask R-CNN (MR), CMR-Cascade Mask R-CNN (CMR), Hybrid Task Cascade without a semantic branch (HTC*), Faster R-CNN (FR), Deformable RoI Pooling (Dp) and RoI Transformer (RT).
				The short names for categories are defined as: {\em BD--Baseball diamond, GTF--Ground field track, SV--Small vehicle,  LV--Large vehicle, TC--Tennis court, BC--Basketball court, SC--Storage tank, SBF--Soccer-ball field, RA--Roundabout, SP--Swimming pool, HC--Helicopter}.}
				\vspace{-3mm}
				\centering
				\setlength{\tabcolsep}{0.9mm}{
					\resizebox{0.95\linewidth}{!}{
						\begin{tabular}{ccccccccccccccccc}
							\hline
							\multicolumn{17}{c}{OBB Results}                                                                                                                  \\ \hline
							method          & Plane & BD    & Bridge & GTF   & SV    & LV    & Ship  & TC    & BC    & ST    & SBF   & RA    & Harbor & SP    & HC    & mAP   \\ \hline
							RetinaNet~\cite{focal}       & 86.54 & 77.45 & 42.8   & 64.87 & 71.06 & 58.5  & 73.53 & 90.72 & 80.97 & 66.67 & 52.42 & 62.16 & 60.79  & 64.84 & 40.83 & 66.28 \\
							MR~\cite{MaskR-CNN}              & 88.7  & 74.13 & 50.75  & 63.66 & 73.64 & 73.98 & 83.68 & 89.74 & 78.92 & 80.26 & 47.43 & 65.09 & 64.79  & 66.09 & 59.79 & 70.71 \\
							CMR~\cite{hybridcascade}             & 88.93 & 75.21 & 51.55  & 64.9  & 74.39 & 75.37 & 84.74 & 90.23 & 77.48 & 81.51 & 46.57 & 63.49 & 65.39  & 67.63 & 56.96 & 70.96 \\
							HTC*~\cite{hybridcascade}            & 89.17 & 75.05 & 51.95  & 64.5  & 74.19 & 76.3  & 86.05 & 90.55 & 79.51 & 77.18 & 50.3  & 61.23 & 65.89  & 68.29 & 58.01 & 71.21 \\
							FR OBB~\cite{DOTA}          & 88.42 & 74.24 & 45.31  & 61.49 & 73.53 & 70.03 & 77.76 & 90.87 & 81.8  & 82.64 & 48.75 & 60.14 & 63.42  & 67.65 & 54.31 & 69.36 \\
							FR OBB + Dp~\cite{Deformable}     & 88.82 & 74.12 & 45.44  & 63.07 & 73.13 & 73.59 & 84.39 & 90.71 & 82.28 & 83.59 & 42.76 & 58.49 & 63.52  & 68.25 & 59.89 & 70.14 \\
							FR H-OBB~\cite{DOTA}        & 88.41 & 79.35 & 45.39  & 63.16 & 73.91 & 72.11 & 83.86 & 90.25 & 77.34 & 81.04 & 48.5  & 60.53 & 63.88  & 66.43 & 57.53 & 70.11 \\
							FR OBB + RT~\cite{RoITransformer}     & 88.34 & 77.07 & 51.63  & 69.62 & 77.45 & 77.15 & 87.11 & 90.75 & 84.9  & 83.14 & 52.95 & 63.75 & 74.45  & 68.82 & 59.24 & 73.76 \\ \hline
							\multicolumn{17}{c}{HBB Results}                                                                                                                  \\ \hline
							method & Plane & BD    & Bridge & GTF   & SV    & LV    & Ship  & TC    & BC    & ST    & SBF   & RA    & Harbor & SP    & HC    & mAP   \\ \hline
							RetinaNet~\cite{focal}       & 88.28 & 77.76 & 47.47  & 59.07 & 73.83 & 63.49 & 77.69 & 90.43 & 78.57 & 65.87 & 48.67 & 61.82 & 68.92  & 71.59 & 38.22 & 67.45 \\
							RetinaNet OBB~\cite{focal} & 88.51 & 78.21 & 47.9   & 64.46 & 75.17 & 72.9  & 78.5  & 90.72 & 82.22 & 67.11 & 51.33 & 62.16 & 69.47  & 69.79 & 37.3  & 69.05 \\
							MR~\cite{MaskR-CNN}              & 88.79 & 79.06 & 53.04  & 63.14 & 78.22 & 65.23 & 77.94 & 89.6  & 81.99 & 81.06 & 47.17 & 65.19 & 72.72  & 68.67 & 62.27 & 71.61 \\
							CMR~\cite{hybridcascade}             & 88.93 & 81.5  & 52.85  & 64.01 & 79.18 & 65.85 & 78.12 & 90.08 & 77.48 & 81.87 & 46.45 & 62.88 & 72.99  & 69.95 & 58.32 & 71.36 \\
							HTC*~\cite{hybridcascade}             & 89.11 & 79.76 & 53.87  & 64.4  & 79.06 & 76.23 & 86.57 & 90.5  & 79.51 & 81.66 & 50.44 & 63.75 & 73.33  & 70.61 & 48.59 & 72.49 \\
							FR~\cite{FasterR-CNN}              & 89.02 & 75.8  & 53.47  & 60.8  & 78.02 & 65.56 & 78.01 & 90.1  & 77.3  & 81.7  & 47.12 & 61.48 & 72.59  & 70.89 & 59.54 & 70.76 \\
							FR OBB~\cite{DOTA}          & 88.37 & 75.45 & 52.11  & 59.98 & 78.08 & 73.42 & 85.65 & 90.81 & 83.22 & 83.18 & 51.45 & 60.01 & 72.44  & 71.09 & 53.44 & 71.91 \\
							FR OBB + Dp~\cite{Deformable}     & 88.86 & 77.4  & 51.6   & 63.12 & 77.62 & 74.73 & 86.2  & 90.72 & 82.73 & 83.75 & 44.35 & 58.81 & 71.97  & 70.98 & 54.68 & 71.83 \\
							FR H-OBB~\cite{DOTA}        & 88.67 & 78.95 & 52.63  & 57.34 & 78.55 & 65.22 & 78.08 & 90.69 & 82.01 & 82.06 & 46.08 & 61.46 & 72.43  & 70.34 & 51.12 & 70.37 \\
							FR OBB + RT~\cite{RoITransformer}     & 88.47 & 81.0  & 54.1   & 69.19 & 78.42 & 81.16 & 87.35 & 90.75 & 84.9  & 83.55 & 52.63 & 62.97 & 75.89  & 71.31 & 57.22 & 74.59 \\ \hline
						\end{tabular}\label{DOTA1cls}
					}
				}\vspace{-3mm}
			\end{table*}
			\begin{table*}[t!]
				\footnotesize
				\caption{Baseline results of class-wise AP on DOTA-v1.5. The abbreviations of algorithms are: Mask R-CNN (MR), CMR-Cascade Mask R-CNN (CMR), Hybrid Task Cascade without a semantic branch (HTC*), Faster R-CNN (FR), Deformable RoI Pooling (Dp) and RoI Transformer (RT). 
				The short names for categories are defined as: {\em BD--Baseball diamond, GTF--Ground field track, SV--Small vehicle,  LV--Large vehicle, TC--Tennis court, BC--Basketball court, SC--Storage tank, SBF--Soccer-ball field, RA--Roundabout, SP--Swimming pool, HC--Helicopter, CC-Container Crane}.}
				\vspace{-3mm}
				\centering				\setlength{\tabcolsep}{0.9mm}{
					\resizebox{0.95\linewidth}{!}{
						\begin{tabular}{cccccccccccccccccc}
							\hline
							\multicolumn{18}{c}{OBB Results} \\ \hline
							method          & Plane & BD    & Bridge & GTF   & SV    & LV    & Ship  & TC    & BC    & ST    & SBF   & RA    & Harbor & SP    & HC    & CC    & mAP   \\ \hline
							RetinaNet~\cite{focal}       & 71.43 & 77.64 & 42.12  & 64.65 & 44.53 & 56.79 & 73.31 & 90.84 & 76.02 & 59.96 & 46.95 & 69.24 & 59.65  & 64.52 & 48.06 & 0.83  & 59.16 \\
							MR~\cite{MaskR-CNN}              & 76.84 & 73.51 & 49.9   & 57.8  & 51.31 & 71.34 & 79.75 & 90.46 & 74.21 & 66.07 & 46.21 & 70.61 & 63.07  & 64.46 & 57.81 & 9.42  & 62.67 \\
							CMR~\cite{hybridcascade}             & 77.77 & 74.62 & 51.09  & 63.44 & 51.64 & 72.9  & 79.99 & 90.35 & 74.9  & 67.58 & 49.54 & 72.85 & 64.19  & 64.88 & 55.87 & 3.02  & 63.41 \\
							HTC*~\cite{hybridcascade}            & 77.8  & 73.67 & 51.4   & 63.99 & 51.54 & 73.31 & 80.31 & 90.48 & 75.12 & 67.34 & 48.51 & 70.63 & 64.84  & 64.48 & 55.87 & 5.15  & 63.4  \\
							FR OBB~\cite{DOTA}          & 71.89 & 74.47 & 44.45  & 59.87 & 51.28 & 68.98 & 79.37 & 90.78 & 77.38 & 67.5  & 47.75 & 69.72 & 61.22  & 65.28 & 60.47 & 1.54  & 62.0  \\
							FR OBB + Dp~\cite{Deformable}     & 71.79 & 73.61 & 44.76  & 61.99 & 51.34 & 70.04 & 79.67 & 90.78 & 76.58 & 67.73 & 44.58 & 70.51 & 61.8   & 65.49 & 64.35 & 0.15  & 62.2  \\
							FR H-OBB~\cite{DOTA}        & 71.57 & 74.71 & 46.39  & 63.4  & 51.54 & 70.11 & 79.09 & 90.63 & 76.81 & 67.4  & 48.66 & 70.9  & 63.1   & 65.67 & 56.66 & 4.55  & 62.57 \\
							FR OBB + RT~\cite{RoITransformer}     & 71.92 & 76.07 & 51.87  & 69.24 & 52.05 & 75.18 & 80.72 & 90.53 & 78.58 & 68.26 & 49.18 & 71.74 & 67.51  & 65.53 & 62.16 & 9.99  & 65.03 \\ \hline
							\multicolumn{18}{c}{HBB Results}                                                                                                                          \\ \hline
							method          & Plane & BD    & Bridge & GTF   & SV    & LV    & Ship  & TC    & BC    & ST    & SBF   & RA    & Harbor & SP    & HC    & CC    & mAP   \\ \hline
							RetinaNet~\cite{focal}       & 74.05 & 77.75 & 48.75  & 59.94 & 49.23 & 61.43 & 77.31 & 90.38 & 75.46 & 61.19 & 47.29 & 69.99 & 67.99  & 74.15 & 40.88 & 10.4  & 61.64 \\
							RetinaNet OBB~\cite{focal} & 71.66 & 77.22 & 48.71  & 65.16 & 49.48 & 69.64 & 79.21 & 90.84 & 77.21 & 61.03 & 47.3  & 68.69 & 67.22  & 74.48 & 46.16 & 5.78  & 62.49 \\
							MR~\cite{MaskR-CNN}              & 78.36 & 77.41 & 53.36  & 56.94 & 52.17 & 63.6  & 79.74 & 90.31 & 74.28 & 66.41 & 45.49 & 71.32 & 70.77  & 73.87 & 61.49 & 17.11 & 64.54 \\
							CMR~\cite{hybridcascade}             & 78.61 & 75.43 & 54.0   & 63.76 & 52.55 & 63.93 & 79.88 & 90.06 & 75.05 & 67.83 & 45.76 & 72.48 & 72.1   & 74.11 & 53.9  & 9.59  & 64.31 \\
							HTC*~\cite{hybridcascade}             & 78.41 & 74.41 & 53.41  & 63.17 & 52.45 & 63.56 & 79.89 & 90.34 & 75.17 & 67.64 & 48.44 & 69.94 & 72.13  & 74.02 & 56.42 & 12.14 & 64.47 \\
							FR~\cite{FasterR-CNN}              & 71.88 & 74.06 & 52.69  & 62.35 & 52.08 & 63.22 & 79.69 & 90.55 & 76.91 & 66.86 & 47.84 & 70.72 & 70.61  & 68.55 & 63.34 & 15.26 & 64.16 \\
							FR OBB~\cite{DOTA}          & 71.91 & 71.6  & 50.58  & 61.95 & 51.99 & 71.05 & 80.16 & 90.78 & 77.16 & 67.66 & 47.93 & 69.35 & 69.51  & 74.4  & 60.33 & 5.17  & 63.85 \\
							FR OBB + Dp~\cite{Deformable}     & 71.9  & 72.8  & 50.84  & 61.99 & 51.97 & 72.15 & 80.13 & 90.74 & 76.53 & 67.95 & 45.09 & 69.93 & 70.43  & 74.72 & 58.28 & 3.19  & 63.67 \\
							FR H-OBB~\cite{DOTA}        & 71.6  & 74.14 & 52.69  & 62.85 & 52.13 & 71.32 & 79.75 & 90.64 & 76.29 & 67.52 & 49.31 & 71.27 & 72.11  & 73.91 & 55.11 & 10.26 & 64.43 \\
							FR OBB + RT~\cite{RoITransformer}     & 71.92 & 75.21 & 54.09  & 68.1  & 52.54 & 74.87 & 80.79 & 90.46 & 78.58 & 68.41 & 51.57 & 71.48 & 74.91  & 74.84 & 56.66 & 13.01 & 66.09 \\ \hline
						\end{tabular}\label{DOTA15cls}
					}
				}\vspace{-3mm}
			\end{table*}
			
			\begin{table*}[t!]
				\footnotesize
				\caption{Baseline results of class-wise AP on DOTA-v2.0. 
				The abbreviations of algorithms are: Mask R-CNN (MR), CMR-Cascade Mask R-CNN (CMR), Hybrid Task Cascade without a semantic branch (HTC*), Faster R-CNN (FR), Deformable RoI Pooling (Dp) and RoI Transformer (RT).
				The short names for categories are defined as: {\em BD--Baseball diamond, GTF--Ground field track, SV--Small vehicle,  LV--Large vehicle, TC--Tennis court, BC--Basketball court, SC--Storage tank, SBF--Soccer-ball field, RA--Roundabout, SP--Swimming pool, HC--Helicopter, CC--Container Crane, Air--Airport, Heli--Helipad}.
				}
				\centering
				\vspace{-3mm}
				\setlength{\tabcolsep}{0.9mm}{
					\resizebox{0.95\linewidth}{!}{
						\begin{tabular}{cccccccccccccccccccc}
							\hline
							\multicolumn{20}{c}{OBB Results}                                                                                                                                      \\ \hline
							method        & Plane & BD    & Bridge & GTF   & SV    & LV    & Ship  & TC    & BC    & ST    & SBF   & RA    & Harbor & SP    & HC    & CC    & Air   & Heli  & mAP   \\ \hline
							RetinaNet~\cite{focal}     & 70.63 & 47.26 & 39.12  & 55.02 & 38.1  & 40.52 & 47.16 & 77.74 & 56.86 & 52.12 & 37.22 & 51.75 & 44.15  & 53.19 & 51.06 & 6.58  & 64.28 & 7.45  & 46.68 \\
							MR~\cite{MaskR-CNN}            & 76.2  & 49.91 & 41.61  & 60.0  & 41.08 & 50.77 & 56.24 & 78.01 & 55.85 & 57.48 & 36.62 & 51.67 & 47.39  & 55.79 & 59.06 & 3.64  & 60.26 & 8.95  & 49.47 \\
							CMR~\cite{hybridcascade}           & 77.01 & 47.54 & 41.79  & 58.02 & 41.58 & 51.74 & 57.86 & 78.2  & 56.75 & 58.5  & 37.89 & 51.23 & 49.38  & 55.98 & 54.59 & 12.31 & 67.33 & 3.01  & 50.04 \\
							HTC*~\cite{hybridcascade}          & 77.69 & 47.25 & 41.15  & 60.71 & 41.77 & 52.79 & 58.87 & 78.74 & 55.22 & 58.49 & 38.57 & 52.48 & 49.58  & 56.18 & 54.09 & 4.2   & 66.38 & 11.92 & 50.34 \\
							FR OBB~\cite{DOTA}        & 71.61 & 47.2  & 39.28  & 58.7  & 35.55 & 48.88 & 51.51 & 78.97 & 58.36 & 58.55 & 36.11 & 51.73 & 43.57  & 55.33 & 57.07 & 3.51  & 52.94 & 2.79  & 47.31 \\
							FR OBB + Dp~\cite{Deformable}   & 71.55 & 49.74 & 40.34  & 60.4  & 40.74 & 50.67 & 56.58 & 79.03 & 58.22 & 58.24 & 34.73 & 51.95 & 44.33  & 55.1  & 53.14 & 7.21  & 59.53 & 6.38  & 48.77 \\
							FR H-OBB~\cite{DOTA}      & 71.39 & 47.59 & 39.82  & 59.01 & 41.51 & 49.88 & 57.17 & 78.36 & 56.87 & 58.24 & 37.66 & 51.86 & 44.61  & 55.49 & 54.74 & 7.56  & 61.88 & 6.6   & 48.9  \\
							FR OBB + RT~\cite{RoITransformer}   & 71.81 & 48.39 & 45.88  & 64.02 & 42.09 & 54.39 & 59.92 & 82.7  & 63.29 & 58.71 & 41.04 & 52.82 & 53.32  & 56.18 & 57.94 & 25.71 & 63.72 & 8.7   & 52.81 \\ \hline
							\multicolumn{20}{c}{HBB Results}                                                                                                                                        \\ \hline
							method        & Plane & BD    & Bridge & GTF   & SV    & LV    & Ship  & TC    & BC    & ST    & SBF   & RA    & Harbor & SP    & HC    & CC    & Air   & Heli  & mAP   \\ \hline
							RetinaNet~\cite{focal}     & 71.86 & 48.69 & 42.2   & 53.12 & 41.16 & 45.64 & 55.9  & 77.74 & 56.14 & 52.0  & 37.68 & 51.46 & 53.27  & 57.51 & 46.76 & 15.66 & 67.76 & 12.97 & 49.31 \\
							RetinaNet OBB~\cite{focal} & 70.99 & 46.77 & 43.76  & 55.08 & 41.55 & 51.06 & 58.01 & 77.78 & 57.72 & 53.5  & 37.66 & 51.72 & 53.56  & 57.32 & 49.18 & 12.01 & 64.53 & 4.48  & 49.26 \\
							MR~\cite{MaskR-CNN}            & 77.61 & 51.35 & 44.89  & 60.12 & 42.51 & 48.1  & 57.93 & 77.84 & 57.55 & 57.88 & 36.53 & 51.71 & 54.79  & 58.93 & 60.01 & 14.42 & 60.32 & 8.43  & 51.16 \\
							CMR~\cite{hybridcascade}           & 78.12 & 47.89 & 46.43  & 57.8  & 42.97 & 48.23 & 59.11 & 78.19 & 57.17 & 58.88 & 37.42 & 51.32 & 53.66  & 58.07 & 55.5  & 17.67 & 67.37 & 1.81  & 50.98 \\
							HTC*~\cite{hybridcascade}          & 78.28 & 47.95 & 45.7   & 59.95 & 42.97 & 48.7  & 59.14 & 78.58 & 55.91 & 58.77 & 37.75 & 52.46 & 53.34  & 58.64 & 55.53 & 9.78  & 66.67 & 5.74  & 50.88 \\
							FR~\cite{FasterR-CNN}            & 76.14 & 49.93 & 44.97  & 57.8  & 42.4  & 47.86 & 57.76 & 77.7  & 56.57 & 58.65 & 39.24 & 52.6  & 54.94  & 58.92 & 56.62 & 12.88 & 61.64 & 6.24  & 50.71 \\
							FR OBB~\cite{DOTA}        & 71.68 & 45.8  & 45.56  & 58.7  & 42.18 & 51.28 & 59.28 & 79.01 & 58.74 & 58.75 & 37.26 & 51.93 & 52.36  & 58.08 & 54.12 & 8.48  & 53.01 & 2.4   & 49.37 \\
							FR OBB + Dp~\cite{Deformable}   & 71.58 & 47.68 & 46.16  & 60.48 & 42.34 & 52.55 & 59.48 & 79.07 & 59.61 & 58.46 & 35.35 & 53.73 & 53.12  & 58.33 & 52.06 & 12.56 & 59.76 & 6.38  & 50.48 \\
							FR H-OBB~\cite{DOTA}      & 77.14 & 50.54 & 45.6   & 57.53 & 42.27 & 48.09 & 57.6  & 78.4  & 59.78 & 57.8  & 36.64 & 52.13 & 52.51  & 58.42 & 48.91 & 14.99 & 60.0  & 8.47  & 50.38 \\
							FR OBB + RT~\cite{RoITransformer}   & 71.84 & 48.2  & 47.84  & 63.94 & 42.97 & 54.79 & 60.74 & 82.88 & 63.51 & 58.89 & 40.63 & 52.83 & 55.7   & 58.87 & 57.94 & 27.04 & 64.27 & 7.68  & 53.37 \\ \hline
						\end{tabular}\label{DOTA2cls}
					}
				}
			\vspace{-3mm}
			\end{table*}
				
\subsection{Codebase and Development Kit}
				We also provide an aerial object detection code library\footnote{\url{https://github.com/dingjiansw101/AerialDetection}} and a development kit\footnote{\url{https://github.com/CAPTAIN-WHU/DOTA_devkit}} for using DOTA. To construct the comprehensive baselines, we select MMDetection as the fundamental code library since it contains rich object detection algorithms and has the feature of modular design. However, the original MMDetection~\cite{mmdetection} lacks the modules to support oriented object detection. Therefore, we enriched MMDetection with {\bf OBB Head} as described in Sec.~\ref{obb_baseline_frame} to enable OBB predictions. We also implemented modules such as rotated RoI Align and rotated position-sensitive RoI Align for rotated region feature extraction, which are crucial components in algorithms such as rotated region proposal network (RRPN)~\cite{RRPN} and RoI Transformer~\cite{RoITransformer}. These new modules are compatible with the modularly designed MMDetection, so we can easily create new algorithms for oriented object detection not restricted to the baseline methods in this paper. We also provide a development kit containing necessary functions for object detection in DOTA, including:
                \begin{itemize}
                    \item[-] {{\em Loading and visualizing the ground truths}.}
                    \item[-] {{\em Calculating the IoU between OBBs}, which is implemented in a mixture of Python/C program. We provide both the CPU and GPU versions.}
                    \item[-] {{\em Evaluating the results.} The evaluation metrics are described in Sec.~\ref{evaluation_metric}}.
                    \item[-] {{\em Cropping and merging images.}} When using the large-size images in DOTA, one can utilize this tool kit to split an original image into patches. 
                    After testing on the patches, one can use the tools to map the results of patches back to the original image coordinates and apply NMS.
                \end{itemize}

				\vspace{-3mm}
				\section{Results}

				\subsection{Benchmark Results and Analyses}\label{sec:benchmarkresults}
				In this section, we conduct a comprehensive evaluation of over 70 experiments and analyze the results. First, we demonstrate the baseline results of 10 algorithms on DOTA-v.10, DOTA-v1.5 and DOTA-v2.0. The baselines cover both two-stage and one-stage algorithms. For most algorithms, we report the mAPs of HBB and OBB predictions, respectively, except for RetinaNet and Faster R-CNN, since they do not support oriented object detection. For algorithms with only OBB heads (RetinaNet OBB, Faster R-CNN OBB, Faster R-CNN OBB +DPool, Faster R-CNN OBB +RoI Transformer), we obtain their HBB results by transferring from OBB as described in Sec.~\ref{sec:horizontal_baseline}. For algorithms with both HBB and OBB heads (Mask R-CNN, Cascade Mask R-CNN, Hybrid Task Cascade*, and Faster R-CNN H-OBB), the HBB mAP is the maximum of the predicted HBB mAP and the transferred HBB mAP. It can be seen that the OBB mAP is usually slightly lower than the HBB mAP for the same algorithm since the OBB task needs a more precise location than the HBB task.
				
\begin{figure}[t!]
					\centering	        	\includegraphics[width=0.98\linewidth]{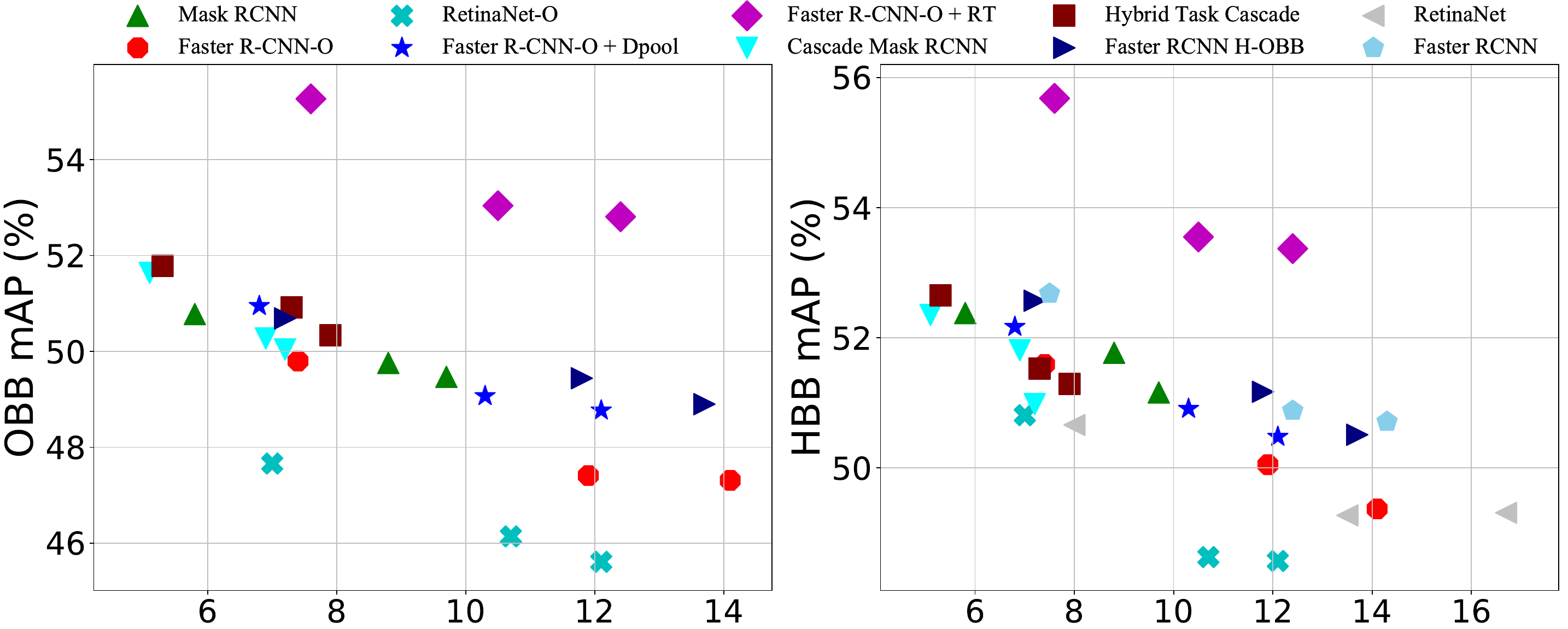}
					\vspace{-2mm}
					\caption{{\bf Results of using different backbones.} The algorithms are tested on DOTA-2.0 test-dev. For each algorithm, we choose 3 different backbones, \ie, ResNet-50 with an FPN, ResNet-101 with an FPN, and 64\(\times\)4d ResNeXt-101 with an FPN. Faster R-CNN-O is the Faster R-CNN OBB in this work. RetinaNet-O stands for RetinaNet OBB. Dpool means the Deformable RoI Pooling, and RT means the RoI Transformer. The speeds are tested on a single Tesla V100.
					}					\label{fig:speed_accuracy}
					\vspace{-4mm}
				\end{figure}

				Tab.~\ref{benchmarks} shows that the performance on DOTA-v1.0, DOTA-v1.5 and DOTA-v2.0 are declining, indicating the increased difficulty of the datasets. 
The class-wise AP are reported in Tab.~\ref{DOTA1cls}, Tab.~\ref{DOTA15cls} and Tab.~\ref{DOTA2cls}. To give more detailed comparisons of speed \textit{vs.} accuracy, we evaluate all algorithms using different backbones (as shown in Fig.~\ref{fig:speed_accuracy}). From the speed-accuracy curve, the Faster R-CNN OBB + RoI Transformer outperforms the other methods. To explore the properties of DOTA and provide guidelines for future research, we evaluate the module design and the hyperparameter setting. Then, we analyze the influence of data augmentation in detail. Finally, we visualize the results to show the difficulties of ODAI.

				\subsubsection{Mask Head \textit{vs.} OBB Head}\label{sec:maskvsobb}
				The OBB head tackles oriented object detection as a regression problem, while the mask head tackles oriented object detection as a pixel-level classification problem.  The mask head more easily converges and achieves better results but is more computationally expensive. Taking the results on the DOTA-v2.0 test-dev set as an example, Mask R-CNN still outperforms Faster R-CNN H-OBB by 0.57 points in OBB mAP. Nevertheless, Mask R-CNN is slower than Faster R-CNN H-OBB by 4 fps. Note that the process of transferring the mask to the OBB is not considered here. Otherwise, Mask R-CNN should be slower.
				
				\subsubsection{RoI Transformer \textit{vs.} Deformable RoI Pooling}
				Geometric variations are still challenging in object detection. In this part, we evaluate RoI Transformer and Dpool by replacing RoI Align in Faster R-CNN OBB. We call these two models Faster R-CNN OBB + RoI Transformer and Faster R-CNN OBB + Dpool. Tab.~\ref{benchmarks} and Fig.~\ref{fig:speed_accuracy} show that Dpool improves the performance of Faster R-CNN OBB at most times, while RoI Transformer performs better than Dpool. This finding verifies that carefully designed geometry transformation modules such as RoI Transformer are better than general geometry transformation modules such as Dpool for aerial images.
				
				\subsubsection{Excluding Small Instances}
				During the training on DOTA-v1.5 and DOTA-v2.0, many extremely small instances will cause numerical instability. For the experiments in DOTA-v1.5 and DOTA-v2.0, we set a threshold to exclude too small instances. We try to explore the influence of different thresholds on DOTA-v2.0. We filter the small instances by two rules: 1) the area of instance is below a certain threshold, and 2) $max(w, h)$ is below a threshold, where the $w$ and $h$ are the width and height, respectively, of the corresponding HBB. The results in Tab.~\ref{excluding_small} show that small instances have little influence on the results.
				
				\begin{table}[htp!]
					\centering
					\caption{Results after excluding extremely small instances by different thresholds in DOTA-v2.0. There are 642,601 instances before filtering.}
					\vspace{-3mm}
\begin{tabular}{c|c|c}
\hline
\# of filtered Instance & Filtering strategy                                 & HBB mAP \\ \hline
99,317              & area $\leq$ 50 and max(w, h) $\leq$ 10 & 51.08   \\
157,287             & area $\leq$ 80 and max(w, h) $\leq$ 10 & 51.35   \\
158,629             & area $\leq$ 80 and max(w, h) $\leq$ 12 & 50.71   \\ \hline
\end{tabular}\label{excluding_small}
\vspace{-3mm}
				\end{table}
				
				\subsubsection{Number of Proposals}
				\label{num_proposals}
				\begin{table*}[t!]
					\centering
					\small
					\caption{Results using different number of proposals on DOTA-v2.0 test-dev. The speeds are tested on a single Tesla V100 GPU. The other settings are the same with those in Tab.~\ref{benchmarks}.}
					\vspace{-3mm}
					\begin{tabular}{c|c|cccccccccc} 
						\hline
						Method                                                                                     & \# of proposals  & 1,000  & 2,000  & 3,000  & 4,000  & 5,000  & 6,000  & 7,000  & 8,000  & 9,000  & 10,000  \\ 
						\hline						\multirow{3}{*}{\begin{tabular}[c]{@{}c@{}}Faster R-CNN OBB\\  + RoI Transformer \end{tabular}} & OBB mAP (\%)~ & 51.72 & 52.81 & 52.81 & 53.24 & 53.29 & 53.51 & 53.70 & \bf{53.94} & 53.93 & 53.92  \\ 
						\cline{2-12}
						& HBB* mAP      & 52.56 & 53.37 & 53.37 & 54.63 & 54.86 & 55.07 & 55.08 & \bf{55.09} & 55.08 & 55.06  \\ 
						\cline{2-12}
						& speed (fps)   & {\bf 14.4}  & 12.4    & 12.2    & 9.1   & 8.7   & 7.8   & 7.5   & 6.5   & 6     & 5.7    \\ 
						\hline
						\multirow{3}{*}{Faster R-CNN OBB}                                                          & OBB mAP (\%)  & 47.10 & 47.31 & 48.03 & 48.09 & 48.32 & 48.35 & 48.48 & \bf{48.49} & 48.49 & 48.49  \\ 
						\cline{2-12}
						& HBB* mAP      & 48.44 & 49.37 & 49.46 & 49.71 & 49.74 & 50.09 & 50.37 & 50.39 & 50.38 & \bf{50.47}  \\ 
						\cline{2-12}
						& speed (fps)   & {\bf 15.8} & 14.1  & 12.5  & 11.9  & 10.9  & 9.9   & 9.3   & 9.1   & 8.4   & 7.8    \\
						\hline
					\end{tabular}\label{table:num_proposals}
					\vspace{-3mm}
				\end{table*}
				The number of proposals is an important hyperparameter in modern detectors. As mentioned before, the possible number of instances in aerial images is quite different from that in natural images. In DOTA, one \(1,024\times1,024\) image may contain more than 1,000 instances. There is no doubt that the parameters that perform well for natural images are not optimal for aerial images. Here we explore the optimal settings for aerial images. As shown in Tab.~\ref{table:num_proposals}, the number of proposals with the highest performance for Faster R-CNN OBB + RoI Transformer is 8,000. For Faster R-CNN OBB, the increase in the mAP slows at approximately 8,000 proposals. Furthermore, from 1,000 to 10,000 proposals, the improvements in Faster R-CNN + RoI Transformer and Faster R-CNN OBB are 2.2 and 1.39 points in mAP, respectively. However, the increased number of proposals bring more computation. 
			Therefore, for the other experiments in this paper, we choose 2,000 proposals. 
				The optimal number of proposals in DOTA is quite a bit larger than that in PASCAL VOC, where 300 is the optimal number. 
				This finding confirms that the difference between aerial and natural images is again significant. 
				
				\subsubsection{Data Augmentation}\label{sec:data_augmentation}
				In this section, we explore the influence of data augmentation in detail. We followed the multi-scale training, testing, and rotation training strategies in~\cite{li2019learning} and further conduct rotation testing. Note that as data augmentation often produces a huge number of patches and will dramatically increase the time complexity of experiments, we conduct our ablation study of data augmentation on DOTA-v1.5, which is similar to DOTA-v2.0 in data distribution while much smaller.
				The model we select is Faster R-CNN OBB + RoI Transformer. We choose R-50-FPN as the backbone and adopt five data augmentation strategies. The first is the high patch overlap. We change the overlap between patches from 200 to 512 since the large instances may be cut off at the edge. The second and third are multi-scale training and testing, respectively. We resize the original images by factors of [0.5, 1.0, 1.5] and then crop the original images into patches of size $1,024\times1,024$. The fourth is the rotation training. For images with roundabouts and storage tanks, we rotate the patches randomly by four angles $[\pi/2, \pi, -\pi/2, -\pi]$. For images with the other categories, we rotate the angle randomly from $[-\pi, \pi]$ continuously during training. We also rotate the images at four angles ($[0, \pi/2, \pi, 3\pi/2]$) during testing.
				 \jian{When performing test time augmentation, the results from images at different angles and scales are merged through the Non-Maximum Suppression (NMS) process in the original image coordinates.}
                As shown in Tab.~\ref{data_augmentation_ablation}, both scale and rotation data augmentations improve the performance of object detection by a large margin, which is consistent with the large scale and orientation variations in DOTA. Furthermore, this baseline model already used a feature pyramid network (FPN) and RoI Transformer. This indicates that the FPN and RoI Transformer do not completely solve the problem of scale and rotation variations, and geometric modeling with CNNs is still an open problem.
				\begin{table}[t!]
					\centering
					\caption{Data augmentation experiments on DOTA-v1.5. \DingJian{Each column in this table indicates an experiment configuration. The first column represents our baseline method without additional data augmentations, while the other columns gradually add augmentation.} We use Faster R-CNN OBB + RoI Transformer as the baseline. High overlap means an overlap of 512 between patches instead of 200 as in Tab.~\ref{benchmarks}.}
					\vspace{-3mm}
					\setlength{\tabcolsep}{1.6mm}{
						\begin{tabular}{c|c|ccccc}
							\hline
							& \multicolumn{1}{l|}{Baseline} & \multicolumn{5}{c}{Data augmentation}                           \\ \hline
							High overlap &                               & \checkmark & \checkmark & \checkmark & \checkmark & \checkmark  \\
							Multi scale Train  &                               &            & \checkmark & \checkmark & \checkmark & \checkmark  \\
							Multi scale Test   &                               &            &            & \checkmark & \checkmark & \checkmark  \\
							Rotation Train     &                               &            &            &            & \checkmark & \checkmark  \\ 
							Rotation Test      &                               &            &            &            &            & \checkmark  \\ \hline
							OBB mAP            & 65.03                         & 67.57      & 69.44      & 73.62      & 76.43      & {\bf 77.60 } \\
							HBB mAP            & 66.09                         & 67.94      & 70.63      & 74.63      & 77.24      & {\bf 78.88 } \\ \hline
						\end{tabular}
					}
					\label{data_augmentation_ablation}
					\vspace{-3mm}
				\end{table}
				
\begin{figure*}
					\centering					\includegraphics[width=0.9\linewidth]{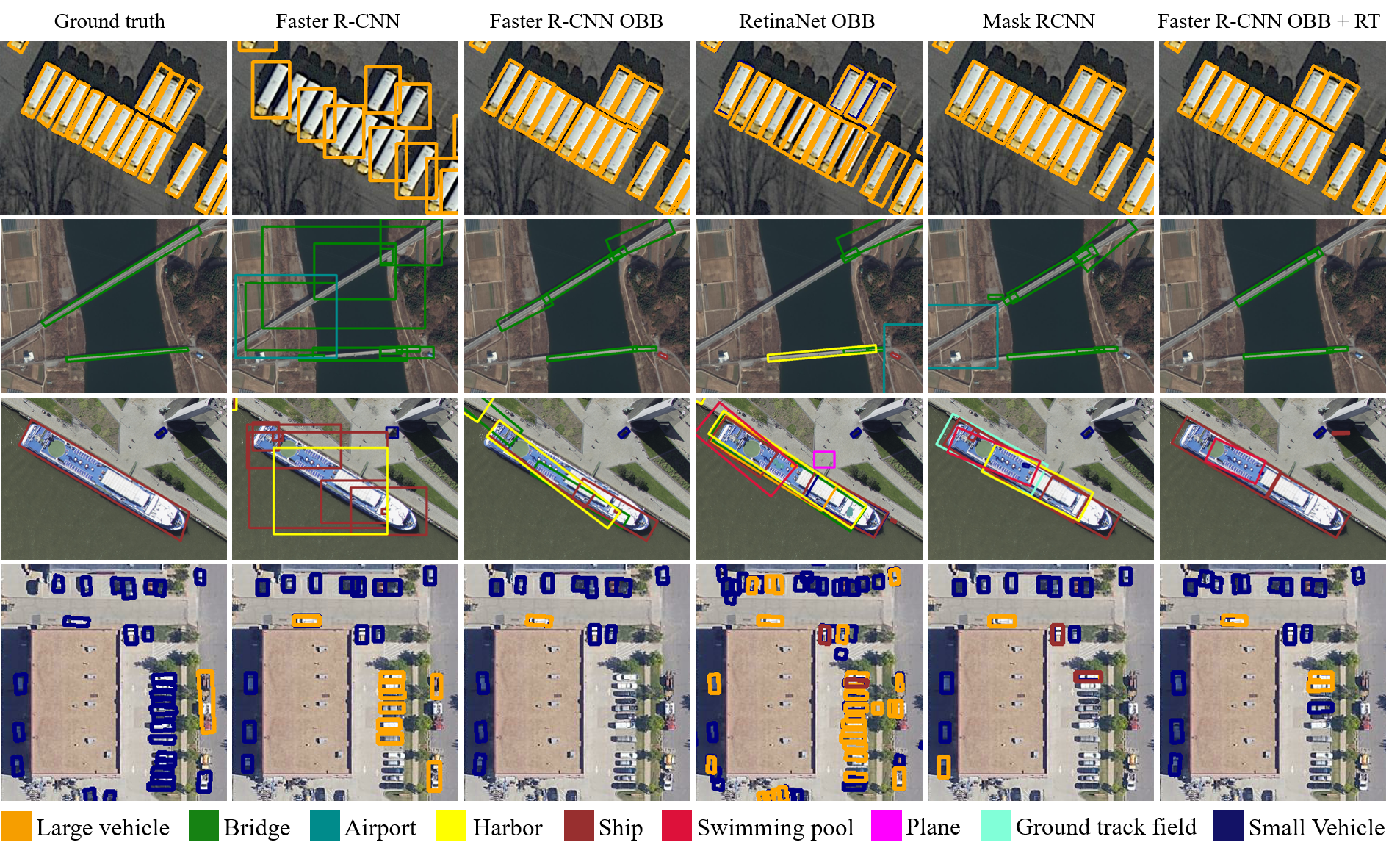}
					\vspace{-2mm}
					\caption{ {\bf Visualization of the results on DOTA-v2.0 test-dev.} The five models are from the DOTA-v2.0 models in Tab.~\ref{benchmarks}. The predictions with scores above 0.1 are shown. The results illustrate the performance in cases of orientation variations, density variations, large ARs and small ARs.}
					\label{fig:vis_comparison}
					\vspace{-4mm}
				\end{figure*}

				\jian{\subsubsection{Class-Wise Results}
				The baseline results of class-wise AP on DOTA-v1.0, DOTA-v1.5, and DOTA-v2.0 are reported in Tab.~\ref{DOTA1cls}, Tab.~\ref{DOTA15cls} and Tab.~\ref{DOTA2cls}. In contrast with  DOTA-v1.0, DOTA-v1.5 additionally annotated the tiny objects (most of them are small vehicles below 10 pixels). Therefore, by comparing the AP of small vehicles of the same detector on DOTA-v1.0 and DOTA-v1.5, we can see the challenges in the detection of tiny objects. Taking Faster R-CNN OBB with RoI Transformer as an example, the AP on small vehicles decrease by \textbf{25.4 points} from \textbf{77.45} to \textbf{52.05}. The challenge on tiny object detection can also be checked in the last row of Fig.~\ref{fig:vis_comparison}. The advantage of OBBs over HBBs in detecting the densely packed objects can be demonstrated by comparing the mAP of OBB detectors and HBB detectors. For example, Faster R-CNN \textbf{OBB} outperforms Faster R-CNN by \textbf{8 points} in AP on large vehicles in DOTA-v1.0. Some examples of the detection of densely packed objects are shown in the first row of Fig.~\ref{fig:vis_comparison}.
				}
				
				\subsubsection{Visualization of the Results}

				We show the performance of Faster R-CNN~\cite{FasterR-CNN}, Faster R-CNN OBB, RetinaNet OBB, Mask R-CNN and Faster R-CNN OBB + RoI Transformer on difficult scenes in Fig.~\ref{fig:vis_comparison}: 1) The first row demonstrates densely packed large vehicles. Faster R-CNN misses many instances due to the high overlaps between neighboring large vehicles in the HBBs. Those large vehicles are suppressed through NMS. Faster R-CNN OBB, Mask R-CNN and Faster R-CNN OBB + RT perform well, while RetinaNet OBB has lower location precision due to feature misalignment. 2) The second and third rows show long shape instances with a large ARs. These instances are self-similar, which means that each part of the instance has a similar feature as the whole instance. For example, the second row shows that all methods have at least two predictions on a single bridge. The third row also reveals this problem. There exist several predictions on a single ship. 3) The second and third rows also indicate that several different categories have very similar features. Bridges are easily classified as airports and harbors while the ships are easily to be classified as harbors and bridges. 4) The last row shows the difficulty of detecting extremely small instances (less than or approximately 10 pixels). The recall of the extremely small instances is very low.
				
				\vspace{-3mm}			\subsection{State-of-the-Art Results on DOTA-v1.0}
				
				\begin{table*}[t!]
					\centering
					\caption{State-of-the-art results on DOTA-v1.0~\cite{DOTA}. The short names for categories are defined as: {\em BD--Baseball diamond, GTF--Ground field track, SV--Small vehicle,  LV--Large vehicle, TC--Tennis court, BC--Basketball court, SC--Storage tank, SBF--Soccer-ball field, RA--Roundabout, SP--Swimming pool, and  HC--Helicopter}. FR-O indicates the {\em Faster R-CNN OBB} detector, which is the previous official baseline provided by DOTA-v1.0~\cite{DOTA}. ICN~\cite{ICN} is the {\em image cascade network}. The LR-O + RT means {\em Light Head R-CNN + RoI Transformer}. DR-101-FPN means {\em deformable ResNet-101 with an FPN}. SCRDet means {\em small, cluttered and rotated object detector}. R-101-SF-MDA means ResNet-101 with sampling fusion network (SF-Net) and multi-dimensional attention network (MDA-Net). RT means {\em RoI Transformer}. Aug. means the data augmentations in Sec.~\ref{sec:data_augmentation}. FR-O* means the re-implemented Faster R-CNN OBB detector, which is slightly different from the original FR-O~\cite{DOTA}.}
					\vspace{-3mm}
					\setlength{\tabcolsep}{0.9mm}{
\resizebox{0.95\linewidth}{!}{
\begin{tabular}{cccccccccccccccccc}
\hline
\multicolumn{18}{c}{OBB Results}                                                                                                                                                                                                                                                  \\ \hline
method                             & backbone     & Plane       & BD          & Bridge      & GTF         & SV          & LV          & Ship        & TC          & BC          & ST          & SBF         & RA          & Harbor      & SP          & HC          & mAP         \\ \hline
FR-O~\cite{DOTA}                   & R-101        & 79.42       & 77.13       & 17.70       & 64.05       & 35.30       & 38.02       & 37.16       & 89.41       & 69.64       & 59.28       & 50.30       & 52.91       & 47.89       & 47.40       & 46.30       & 54.13       \\
ICN~\cite{ICN}                     & DR-101-FPN   & 81.36       & 74.30       & 47.70       & 70.32       & 64.89       & 67.82       & 69.98       & 90.76       & 79.06       & 78.20       & 53.64       & 62.90       & 67.02       & 64.17       & 50.23       & 68.16       \\
LR-O + RT~\cite{RoITransformer}    & R-101-FPN    & 88.64       & 78.52       & 43.44       & 75.92       & 68.81       & 73.68       & 83.59       & 90.74       & 77.27       & 81.46       & 58.39       & 53.54       & 62.83       & 58.93       & 47.67       & 69.56       \\
SCRDet~\cite{scrdet}               & R-101-SF-MDA & 89.98       & 80.65       & 52.09       & 68.36       & 68.36       & 60.32       & 72.41       & 90.85       & {\bf 87.94} & 86.86       & 65.02       & 66.68       & 66.25       & 68.24       & 65.21       & 72.61       \\
DRN~\cite{DRN}                     & H-104        & 89.71       & 82.34       & 47.22       & 64.10       & 76.22       & 74.43       & 85.84       & 90.57       & 86.18       & 84.89       & 57.65       & 61.93       & 69.30       & 69.63       & 58.48       & 73.23       \\
Gliding Vertex~\cite{Glidingertex} & R-101-FPN    & 89.64       & 85.00       & 52.26       & 77.34       & 73.01       & 73.14       & 86.82       & 90.74       & 79.02       & 86.81       & 59.55       & {\bf 70.91}   & 72.94       & 70.86       & 57.32       & 75.02       \\
CenterMap~\cite{CenterMap}         & R-101-FPN    & 89.83       & 84.41       & 54.60       & 70.25       & 77.66       & 78.32       & 87.19       & 90.66       & 84.89       & 85.27       & 56.46       & 69.23       & 74.13       & 71.56       & 66.06       & 76.03       \\
CSL~\cite{CSL}                     & R-152-FPN    & 90.25       & 85.53       & 54.64       & 75.31       & 70.44       & 73.51       & 77.62       & 90.84       & 86.15       & 86.69       & 69.60       & 68.04       & 73.83       & 71.10       & 68.93       & 76.17       \\
Li et al.~\cite{li2019learning}    & R-101-FPN    & {\bf 90.41} & {\bf 85.21} & 55.00       & 78.27       & 76.19       & 72.19       & 82.14       & 90.70       & 87.22       & 86.87 & 66.62       & 68.43       & 75.43       & 72.70       & 57.99       & 76.36       \\
S$^2$A-Net~\cite{han2020align}     & R-50-FPN     & 88.89       & 83.60       & 57.74       & {\bf 81.95}   & {\bf 79.94}     & 83.19       & {\bf 89.11}        & 90.78       & 84.87       & {\bf 87.81}      & 70.30       & 68.25       & 78.30       & 77.01       & 69.58       & 79.42       \\
FR-O* + RT~\cite{RoITransformer}                         & R-50-FPN     & 88.34       & 77.07       & 51.63       & 69.62       & 77.45 & 77.15       & 87.11       & 90.75       & 84.90       & 83.14       & 52.95       & 63.75       & 74.45       & 68.82       & 59.24       & 73.76       \\
FR-O* + RT (Aug.)~\cite{RoITransformer}                  & R-50-FPN     & 87.89       & 85.01       & {\bf 57.83} & 78.55 & 75.22       & {\bf 84.37} & 88.04 & {\bf 90.88} & 87.28       & 85.79       & {\bf 71.04} & 69.67 & {\bf 79.00} & {\bf 83.29} & {\bf 73.43} & {\bf 79.82} \\ \hline
\multicolumn{18}{c}{HBB Results}                                                                                                                                                                                                                                                  \\ \hline
method                             & backbone     & Plane       & BD          & Bridge      & GTF         & SV          & LV          & Ship        & TC          & BC          & ST          & SBF         & RA          & Harbor      & SP          & HC          & mAP         \\ \hline
ICN~\cite{ICN}                     & DR-101-FPN   & 89.97       & 77.71       & 53.38       & 73.26       & 73.46       & 65.02       & 78.22       & 90.79       & 79.05       & 84.81       & 57.20       & 62.11       & 73.45       & 70.22       & 58.08       & 72.45       \\
SCRDet~\cite{scrdet}               & R-101-SF-MDA & 90.18       & 81.88       & 55.30       & 73.29       & 72.09       & 77.65       & 78.06       & {\bf 90.91} & 82.44       & 86.39       & 64.53       & 63.45       & 75.77       & 78.21       & 60.11       & 75.35       \\
CenterMap~\cite{CSL}               & R-101-FPN    & 89.70       & 84.92       & 59.72       & 67.96       & {\bf 79.16}      & 80.66       & 86.61       & 90.47       & 84.47       & 86.19       & 56.42       & 69.00       & 79.33       & 80.53       & 64.81       & 77.33       \\
Li et al.~\cite{li2019learning}    & ResNet101    & {\bf 90.41} & {\bf 85.77} & 61.94       & {\bf 78.18} & 77.00       & 79.94       & 84.03       & 90.88       & {\bf 87.30} & {\bf 86.92} & 67.78       & 68.76       & 82.10       & 80.44       & 60.43       & 78.79       \\
FR-O* + RT~\cite{RoITransformer}                         & R-50-FPN     & 88.47       & 81.00       & 54.10       & 69.19       & 78.42 & 81.16       & 87.35       & 90.75       & 84.90       & 83.55       & 52.63       & 62.97       & 75.89       & 71.31       & 57.22       & 74.59       \\
FR-O* + RT (Aug.)~\cite{RoITransformer}                  & R-50-FPN     & 87.91       & 85.11       & {\bf 62.65} & 77.73       & 75.83       & {\bf 85.03} & {\bf 88.18} & 90.88       & 87.28       & 86.18       & {\bf 71.49} & {\bf 70.37} & {\bf 84.94} & {\bf 84.11} & {\bf 73.61} & {\bf 80.75} \\ \hline
\end{tabular}
}
}\label{state-of-art}
\vspace{-3mm}
\end{table*}
\begin{table*}[!h]
\centering
\caption{{\bf DOAI 2019 Challenge Results}. CC is the {\em container crane} for short. The other abbreviations for categories are the same as those in Tab.~\ref{state-of-art}. The USTC-NELSLIP, pca\_lab and czh, AICyber are the top 3 participants in the OBB and HBB Tasks. The FR-O means Faster R-CNN OBB. RT means the RoI Transformer. Aug. means the data augmentation method described in Sec.~\ref{sec:data_augmentation}. Note that FR-O + RT and FR-O + RT (Aug.) are single models, while others are ensembles of multiple models.}
\vspace{-3mm}
\setlength{\tabcolsep}{0.9mm}{
\resizebox{0.95\linewidth}{!}{
\begin{tabular}{cccccccccccccccccc}
\hline 
\multicolumn{18}{c}{OBB results}                                                                                                                           \\ \hline 
team (method)    & Plane & BD    & Bridge & GTF   & SV    & LV    & Ship  & TC    & BC    & ST    & SBF   & RA    & Harbor & SP    & HC    & CC    & mAP   \\ \hline
USTC-NELSLIP~\cite{APE}     & {\bf 89.19} & {\bf 85.32} & 57.27  & 80.86 & {\bf 73.87} & 81.26 & 89.5  & 90.84 & {\bf 85.94} & {\bf 85.62} & 69.5  & 76.73 & 76.34  & 76    & {\bf 77.84} & {\bf 57.33} & {\bf 78.34} \\
pca\_lab~\cite{li2019learning}         & 89.11 & 83.83 & 59.55  & {\bf 82.8} & 66.93 & 82.51 & 89.78 & {\bf 90.88} & 85.36 & 84.22 & 71.95 & {\bf 77.89} & 78.47  & 74.27 & 74.77 & 53.22 & 77.84 \\
czh              & 89    & 83.22 & 54.47  & 73.79 & 72.61 & 80.28 & 89.32 & 90.83 & 84.36 & 85    & 68.68 & 75.3  & 74.22  & 74.41 & 73.45 & 42.13 & 75.69 \\
FR-O + RT~\cite{RoITransformer}        & 71.92 & 76.07 & 51.87  & 69.24 & 52.05 & 75.18 & 80.72 & 90.53 & 78.58 & 68.26 & 49.18 & 71.74 & 67.51  & 65.53 & 62.16 & 9.99  & 65.03 \\
FR-O + RT (Aug.)~\cite{RoITransformer} & 87.54 & 84.34 & {\bf 62.22} & 79.77 & 67.29 & {\bf 83.16} & {\bf 89.93} & 90.86 & 83.85 & 77.74 & {\bf 73.91} & 75.31 & {\bf 78.61} & {\bf 77.07} & 75.20 & 54.77 & 77.60 \\ \hline 
\multicolumn{18}{c}{HBB results}                                                                                                                           \\ \hline
team (method)    & Plane & BD    & Bridge & GTF   & SV    & LV    & Ship  & TC    & BC    & ST    & SBF   & RA    & Harbor & SP    & HC    & CC    & mAP   \\ \hline
pca\_lab~\cite{li2019learning}         & 88.26 & {\bf 86.55} & {\bf 65.68}  & 79.83 & 74.59 & 79.35 & 88.12 & {\bf 90.86} & 85.45 & 84.15 & {\bf 73.9}  & {\bf 77.44} & 84.1   & 81.07 & 76.07 & 57.07 & {\bf 79.53} \\
USTC-NELSLIP~\cite{APE}     & {\bf 89.26} & 85.6  & 59.61  & {\bf 80.86} & 75.2  & 81.13 & 89.58 & 90.84 & {\bf 85.94} & 85.71 & 69.5  & 76.34 & 81.7   & 81.84 & {\bf 76.53} & 57.09 & 79.17 \\
AICyber          & 89.2  & 85.56 & 64.44  & 74.07 & {\bf 77.45} & 81.5  & 89.65 & 90.83 & 85.72 & {\bf 86.03} & 69.82 & 76.34 & 82.89  & {\bf 82.95} & 74.64 & 44.02 & 78.44 \\
FR-O + RT~\cite{RoITransformer}          & 71.92 & 75.21 & 54.09  & 68.10 & 52.54 & 74.87 & 80.79 & 90.46 & 78.58 & 68.41 & 51.57 & 71.48 & 74.91  & 74.84 & 56.66 & 13.01 & 66.09 \\
FR-O + RT (Aug.)~\cite{RoITransformer}   & 87.79 & 84.33 & 63.75  & 79.13 & 72.92 & {\bf 83.08} & {\bf 90.04} & {\bf 90.86} & 83.85 & 77.80 & 73.30 & 75.66 & {\bf 84.84}  & 82.16 & 75.20 & {\bf 57.39} & 78.88 \\ \hline
\end{tabular} 
}
\label{doai_2019_results}}
\vspace{-3mm}
\end{table*}				
In this section, we compare the performance of Faster R-CNN OBB + RoI Transformer with the state-of-the-art algorithms on DOTA-v1.0~\cite{DOTA}. As shown in Tab.~\ref{state-of-art},  Faster R-CNN OBB + RoI Transformer achieves an OBB mAP of 73.76 for DOTA-v1.0, and it outperforms all the previous state-of-the-art methods except that proposed by Li et al.~\cite{li2019learning}. Note that the method of Li et al.~\cite{li2019learning}, SCRDet~\cite{scrdet} and the image cascade network (ICN)~\cite{ICN} all use multiple scales for training and testing to achieve high performance. The method in~\cite{li2019learning} further used rotation data augmentation during training as described in Sec.~\ref{sec:data_augmentation}.
When using the same data augmentation, we achieve an mAP of 79.82. It outperforms the method in~\cite{li2019learning} by 3.46 points in OBB mAP and 1.96 points in HBB mAP.
				In addition, there is a significant improvement in densely packed small instances. (\eg, the small vehicles, large vehicles, and ships). For example, the detection performance for the large vehicle category gains an improvement of 12.18 points compared to the previous results.
				
\subsection{DOAI 2019 Challenge Results}

				DOTA-v1.5 was first used to hold the DOAI Challenge-2019 in conjunction with CVPR 2019\footnote{\url{https://captain-whu.github.io/DOAI2019/}}. There were 173 registrations in total, 13 teams submitted valid results on the OBB Task, and 22 teams submitted valid results on the HBB Task. 
				The detailed leaderboards for the two tasks can be found on the DOAI Challenge-2019 website\footnote{\url{https://captain-whu.github.io/DOAI2019/results.html}}, and the top 3 results are listed in Tab.~\ref{doai_2019_results}. 
				Notice that most of these results have been achieved by using an ensemble of detection models, except~\cite{li2019learning} which used a single model and reported $74.9$ and $77.9$ in terms of mAP on the OBB and HBB tasks, respectively. 
				Both in the training and testing phase, multi-scaling and rotation strategies were used for data augmentations. With the same settings, our single model~\cite{RoITransformer} achieved 76.43 and 77.24 in terms of mAP on the OBB and HBB tasks respectively, as shown in Tab.~\ref{data_augmentation_ablation}, which was the best results reported on the OBB task.

				\section{Conclusion}
				ODAI is challenging. To advance future research, we introduce a large-scale dataset, DOTA, containing 1,793,658 instances annotated by OBBs. The DOTA statistics show that it can well represent the real world well. Then, we build a code library for both oriented and horizontal ODAI to conduct a comprehensive evaluation. We hope these experiments can act as benchmarks for fair comparisons between ODAI algorithms. The results show that hyperparameter selection and module design of the algorithms (\eg, number of proposals) for aerial images are very different from those for natural images. It indicates that DOTA can be used as a supplement to natural scene images to facilitate universal object detection.

				In the future, we will continue to extend the dataset, host more challenges, and integrate more algorithms for oriented object detection into our code library. We believe that DOTA, challenges and code library will not only promote the development of object detection in Earth vision but also pose interesting algorithmic questions for general object detection in computer vision.

				\vspace{-2mm}
				\section*{Acknowledgment}
				We thank the the support of CycloMedia B.V. for providing the airborne images in DOTA-v2.0.  
				We thank Huan Yi, Zhipeng Lin, Fan Hu, Pu Jin, Xinyi Tong, Xuan Hu, Zhipeng Dong, Liang Wu, Jun Tang, Linyan Cui, Duoyou Zhou, Tengteng Huang, and all the others who involved in the annotations of DOTA. We also thank Zhen Zhu for his advice on running Faster R-CNN, and Jinwang Wang for valuable discussions in details of parameter settings. The numerical calculations in this paper have been done on the supercomputing system in the Supercomputing Center of Wuhan University.

				\ifCLASSOPTIONcaptionsoff
				\newpage
				\fi

				
				
				%
				
				{\small
					\bibliographystyle{IEEEtran}
					\bibliography{egbib}
				}
				
				
				
				
				%
				\vspace{-16mm}

				\ifCLASSOPTIONcompsoc	
				
\vskip .5\baselineskip plus -1fil

\begin{IEEEbiography}[{\includegraphics[width=1in,height=1.25in,clip,keepaspectratio]{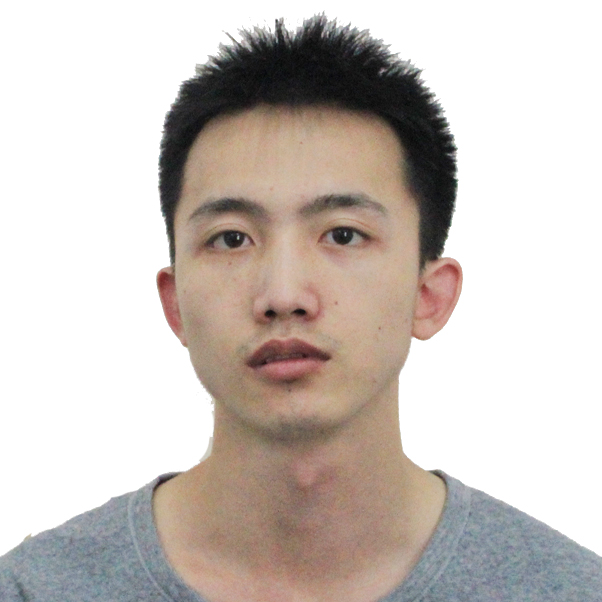}}]{Jian Ding} is currently pursuing his Ph.D degree at the State Key Laboratory of Information Engineering in Surveying, Mapping and Remote Sensing, Wuhan University. He received the B.S. degree in Aircraft Design and Engineering from Northwestern Polytechnical University, Xian, China in 2017. His research interests include object detection, instance segmentation and remote sensing.
\end{IEEEbiography}
				
\vskip -2.6\baselineskip plus -1fil

\begin{IEEEbiography}[{\includegraphics[width=1in,height=1.25in,clip,keepaspectratio]{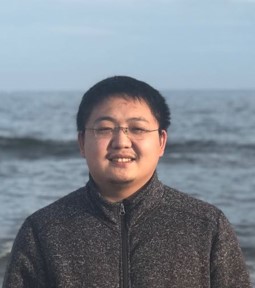}}]{Nan Xue} is currently a Research Associate Professor in the School of Computer Science, Wuhan University. He received the B.S., and Ph.D. degrees from Wuhan University in 2014 and 2020 respectively. He was a visiting scholar at North Carolina State University from Sep. 2018 to June 2020. His research interests include geometric structure analysis in computer vision.
\end{IEEEbiography}
				
\vskip -2.6\baselineskip plus -1fil

\begin{IEEEbiography}[{\includegraphics[width=1in,height=1.25in,clip,keepaspectratio]{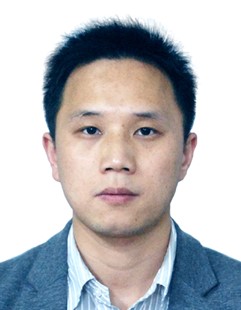}}]{Gui-Song Xia}
received his Ph.D. degree in image processing and computer vision from CNRS LTCI, T{\'e}l{\'e}com ParisTech, Paris, France, in 2011. From 2011 to 2012, he has been a Post-Doctoral Researcher with the Centre de Recherche en Math{\'e}matiques de la Decision, CNRS, Paris-Dauphine University, Paris, for one and a half years.
He is currently working as a full professor 
at Wuhan University. He has also been working as Visiting Scholar at DMA, {\'E}cole Normale Sup{\'e}rieure (ENS-Paris) for two months in 2018. 
His current research interests include mathematical modeling of images and videos, structure from motion, perceptual grouping, and remote sensing image understanding. He serves on the Editorial Boards of several journals, including Pattern Recognition, Signal Processing: Image Communications, EURASIP Journal on Image \& Video Processing, Journal of Remote Sensing, and Frontiers in Computer Science: Computer Vision.
\end{IEEEbiography}

\vskip -2.6\baselineskip plus -1fil

\begin{IEEEbiography}[{\includegraphics[width=1in,height=1.25in,clip,keepaspectratio]{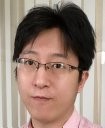}}]{Xiang Bai} received the B.S., M.S., and Ph.D. degrees from the Huazhong University of
Science and Technology (HUST), Wuhan, China, in
2003, 2005, and 2009, respectively, all in electronics
and information engineering.
He is currently a Professor with the School of
Electronic Information and Communications and
the Vice-Director of the National Center of AntiCounterfeiting
Technology, HUST. His research
interests include object recognition, shape analysis,
and scene text recognition.
\end{IEEEbiography}
				
\vskip 1.3\baselineskip plus -1fil

\begin{IEEEbiography}[{\includegraphics[width=1in,height=1.25in,clip,keepaspectratio]{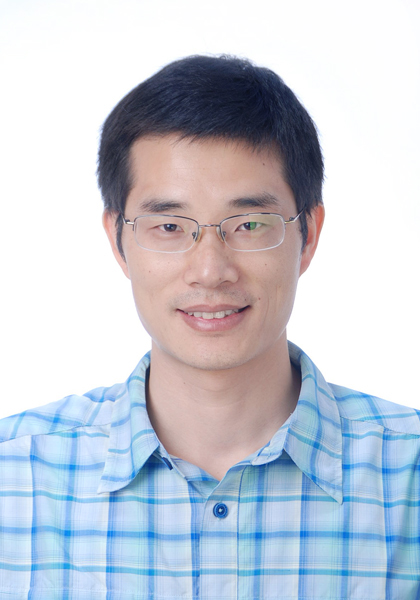}}]{Wen Yang} received the Ph.D. degree in communication and information system from Wuhan University, Wuhan, China, in  2004. From 2008 to 2009, he was a visiting Scholar with the Laboratoire Jean Kuntzmann (LJK), Grenoble, France. From 2010 to 2013, he was a Post-Doctoral Researcher with the State Key Laboratory of Information Engineering, Surveying, Mapping and Remote Sensing (LIESMARS), Wuhan University. Since then, he has been a Full Professor with the School of Electronic Information, Wuhan University. His research interests include object detection and recognition,  semantic segmentation, and change detection.
\end{IEEEbiography}				
				
\vskip -1.3\baselineskip plus -1fil

\begin{IEEEbiography}[{\includegraphics[width=1in,height=1.25in,clip,keepaspectratio]{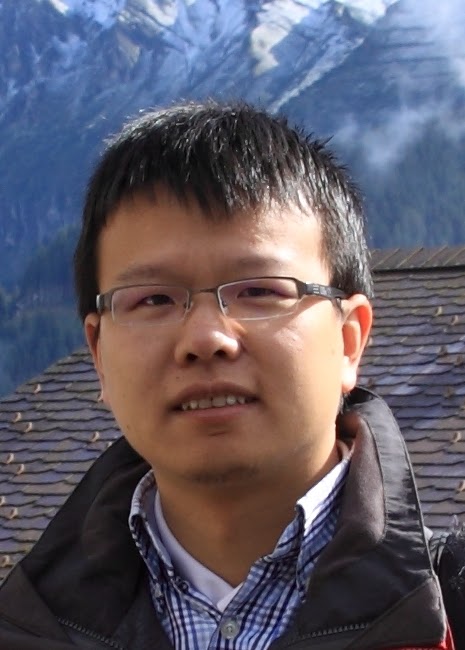}}]{Michael Ying Yang} is currently Assistant Professor in the Department of Earth Observation Science at ITC - Faculty of Geo-Information Science and Earth Observation, University of Twente, The Netherlands, heading a group working on scene understanding.
He received the PhD degree (summa cum laude) from University of Bonn (Germany) in 2011. 
He received the venia legendi in Computer Science from Leibniz University Hannover in 2016.
His research interests are in the fields of computer vision and photogrammetry with specialization on scene understanding and semantic interpretation from imagery. 
He serves as Associate Editor of ISPRS Journal of Photogrammetry and Remote Sensing, Co-chair of ISPRS working group II/5 Dynamic Scene Analysis, 
Program Chair of ISPRS Geospatial Week 2019,
and recipient of ISPRS President's Honorary Citation (2016), Best Science Paper Award at BMVC (2016), and The Willem Schermerhorn Award (2020). 
\end{IEEEbiography}
	
\vskip -2.6\baselineskip plus -1fil

\begin{IEEEbiography}[{\includegraphics[width=1in,height=1.25in,clip,keepaspectratio]{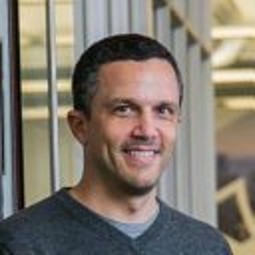}}]{Serge Belongie}
received a B.S. (with honor) in EE from Caltech in 1995 and a Ph.D. in EECS from Berkeley in 2000. While at Berkeley, his research was supported by an NSF Graduate Research Fellowship. From 2001-2013 he was a professor in the Department of Computer Science and Engineering at University of California, San Diego. He is currently a professor at Cornell Tech and the Department of Computer Science at Cornell University. His research interests include Computer Vision, Machine Learning, Crowdsourcing and Human-in-the-Loop Computing. He is also a co-founder of several companies including Digital Persona, Anchovi Labs and Orpix. He is a recipient of the NSF CAREER Award, the
Alfred P. Sloan Research Fellowship, the MIT Technology Review “Innovators Under 35” Award and the Helmholtz Prize for fundamental contributions in Computer Vision.
\end{IEEEbiography}

\vskip -2.6\baselineskip plus -1fil

\begin{IEEEbiography}[{\includegraphics[width=1in,height=1.25in,clip,keepaspectratio]{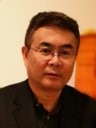}}]{Jiebo Luo}
joined the University of Rochester in Fall 2011 after over fifteen prolific years at Kodak Research Laboratories, where he was a Senior Principal Scientist leading research and advanced development. He has been involved in numerous technical conferences, including serving as the program co-chair of ACM Multimedia 2010, IEEE CVPR 2012 and IEEE ICIP 2017. He has served on the editorial boards of the IEEE Transactions on Pattern Analysis and Machine Intelligence, IEEE Transactions on Multimedia, IEEE Transactions on Circuits and Systems for Video Technology, ACM Transactions on Intelligent Systems and Technology, Pattern Recognition, Machine Vision and Applications, and Journal of Electronic Imaging.  Dr. Luo will serve as the Editor-in-Chief of the IEEE Transactions on Multimedia for the 2020-2022 term. Dr. Luo is a Fellow of SPIE, IAPR, IEEE, ACM, and AAAI. In addition, he is a Board Member of the Greater Rochester Data Science Industry Consortium.
\end{IEEEbiography}

\vskip -2.6\baselineskip plus -1fil

\begin{IEEEbiography}[{\includegraphics[width=1in,height=1.25in,clip,keepaspectratio]{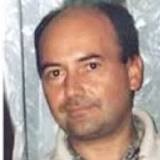}}]{Mihai Datcu} received the M.S. and Ph.D. degrees
in electronics and telecommunications from the University
“Politehnica” of Bucharest UPB, Bucharest,
Romania, in 1978 and 1986, and the title “Habilitation
a diriger des recherches” from Université Louis
Pasteur, Strasbourg, France.
He holds a Professorship in electronics and
telecommunications with UPB since 1981. Since
1993, he has been a Scientist with the German
Aerospace Center (DLR), Oberpfaffenhofen,
Germany. He is currently developing algorithms
for model-based information retrieval from high-complexity signals, methods for scene understanding from SAR and interferometric SAR data, and he is engaged in research in information theoretical aspects and semantic representations
in advanced communication systems. His research interests are in Bayesian inference, information and complexity
theory, stochastic processes, model-based scene understanding, image
information mining, for applications in information retrieval and understanding
of high-resolution SAR and optical observations.
\end{IEEEbiography}

\vskip -2.6\baselineskip plus -1fil

\begin{IEEEbiography}[{\includegraphics[width=1in,height=1.25in,clip,keepaspectratio]{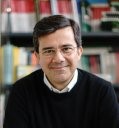}}]{Marcello Pellilo}
is a Full Professor of Computer Science at Ca’ Foscari University,
Venice, where he leads the Computer Vision and Pattern Recognition Lab. He has been the Director of the European Centre for Living
Technology (ECLT) and has held visiting research/teaching positions in several
institutions including Yale University (USA), University College London (UK), McGill
University (Canada), University of Vienna (Austria), York University (UK), NICTA
(Australia), Wuhan University (China), Huazhong University of Science and Technology
(China), and South China University of Technology (China). He is also
an external affiliate of the Computer Science Department at Drexel University (USA). 
His research interests are in the areas of computer vision, machine learning and pattern recognition where he has published more than 200 technical papers in refereed journals, handbooks, and conference proceedings.
He has been General Chair for ICCV 2017, Program Chair for ICPR 2020, and has been Track or Area Chair for several conferences in his area. 
He is the Specialty Chief Editor of Frontiers in Computer Vision and serves, or has served, on the Editorial Boards of several journals, including IEEE Transactions on Pattern Analysis and Machine Intelligence, Pattern Recognition, IET Computer Vision, and Brain Informatics.
He also serves on the Advisory Board of Springer’s International Journal of Machine Learning and Cybernetics.
Prof. Pelillo has been elected Fellow of the IEEE and Fellow of the IAPR, and is an IEEE SMC Distinguished Lecturer.\end{IEEEbiography}

\vskip -2.6\baselineskip plus -1fil

\begin{IEEEbiography}[{\includegraphics[width=1in,height=1.25in,clip,keepaspectratio]{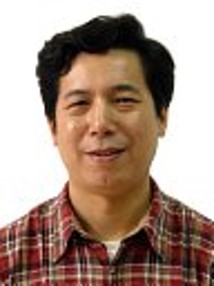}}]{Liangpei Zhang}
received the B.S. degree in physics from Hunan Normal University, Changsha, China, in 1982, the M.S. degree in optics from the Xian Institute of Optics and Precision Mechanics, Chinese Academy of Sciences, Xian, China, in 1988, and the Ph.D. degree in photogrammetry and remote sensing from Wuhan University, Wuhan, China, in 1998. He is currently a Chang-Jiang Scholar Chair Professor with Wuhan University, appointed by the Ministry of Education of China. He has authored or coauthored over 500 research papers and five books. He holds 15 patents. His research interests include hyper spectral remote sensing, high resolution remote sensing, image processing, and artificial intelligence. Dr. Zhang was a recipient of the 2010 Best Paper Boeing Award and the 2013 Best Paper ERDAS Award from the American Society of Photogrammetry and Remote Sensing. He serves as a Co-Chair for the series SPIE Conferences on Multispectral Image Processing and Pattern Recognition, the Conference on Asia Remote Sensing, and many other conferences. He serves as an Associate Editor for the IEEE Transactions on Geoscience and Remote Sensing. He is a fellow of IEEE.
\end{IEEEbiography}
				\fi
				
				
				
				
				

\end{document}